\documentclass[sigconf,screen]{acmart}

\usepackage[utf8]{inputenc} % allow utf-8 input
\usepackage[T1]{fontenc}    % use 8-bit T1 fonts
\usepackage{booktabs}       % professional-quality tables
\usepackage{nicefrac}       % compact symbols for 1/2, etc.
\usepackage{microtype}      % microtypography
\usepackage{xcolor}         % colors
\usepackage{breqn}
\usepackage{bbm}
\usepackage{xspace}
\usepackage{graphicx}
\usepackage{float}
\usepackage{subcaption}
\def\xHyphenate#1#2\wholeString {\if#1$%
    \else\transform{#1}%
    \takeTheRest#2\ofTheString\fi}

\def\takeTheRest#1\ofTheString\fi
{\fi \xHyphenate#1\wholeString}

\def\transform#1{\url{#1}\hskip 0pt plus 1pt}

\newcommand{\fulltitle}{Subverting Fair Image Search with Generative Adversarial Perturbations}

\definecolor{linkColor}{RGB}{6,125,233}
\hypersetup{%
  pdftitle={\fulltitle},
  pdfauthor={},
  pdfkeywords={},
  bookmarksnumbered,
  pdfstartview={FitH},
  colorlinks,
  citecolor=black,
  filecolor=black,
  linkcolor=black,
  urlcolor=linkColor,
  breaklinks=true,
  hypertexnames=false
}

\newtheorem{definition}{Definition}

\newcommand{\eg}{e.g.,\ }
\newcommand{\etal}{et al.\xspace}
\newcommand{\ie}{i.e.,\ }


\copyrightyear{2022}
\acmYear{2022}
\setcopyright{rightsretained}
\acmConference[FAccT '22]{2022 ACM Conference on Fairness, Accountability, and Transparency}{June 21--24, 2022}{Seoul, Republic of Korea}
\acmBooktitle{2022 ACM Conference on Fairness, Accountability, and Transparency (FAccT '22), June 21--24, 2022, Seoul, Republic of Korea}
\acmDOI{10.1145/3531146.3533128}
\acmISBN{978-1-4503-9352-2/22/06}
\begin{document}
\title{\fulltitle}

% The \author macro works with any number of authors. There are two commands
% used to separate the names and addresses of multiple authors: \And and \AND.
%
% Using \And between authors leaves it to LaTeX to determine where to break the
% lines. Using \AND forces a line break at that point. So, if LaTeX puts 3 of 4
% authors names on the first line, and the last on the second line, try using
% \AND instead of \And before the third author name.

\author{Avijit Ghosh}
\email{ghosh.a@northeastern.edu}
\orcid{0000-0002-8540-3698}
\affiliation{%
  \institution{Northeastern University}
  \city{Boston}
  \state{MA}
  \country{USA}
}

\author{Matthew Jagielski}
\email{jagielski@google.com}
\orcid{0000-0002-9749-0696}
\affiliation{%
  \institution{Google Research}
  \city{Mountain View}
  \state{CA}
  \country{USA}
}

\author{Christo Wilson}
\email{cbw@ccs.neu.edu}
\orcid{0000-0002-5268-004X}
\affiliation{%
  \institution{Northeastern University}
  \city{Boston}
  \state{MA}
  \country{USA}
}

\iffalse
\author{%
  Avijit Ghosh \\
  Northeastern University\\
  \texttt{avijit@ccs.neu.edu} \\
  \And
  Matthew Jagielski \\
  Northeastern University\\
  \texttt{jagielski.m@northeastern.edu} \\
  % examples of more authors
   \And
   Christo Wilson \\
   Northeastern University \\
  % Address \\
   \texttt{cbw@ccs.neu.edu} \\
}
\fi


\makeatletter
\renewcommand{\sectionautorefname}{\S\,\@gobble}
\renewcommand{\subsectionautorefname}{\S\,\@gobble}
\renewcommand{\subsubsectionautorefname}{\S\,\@gobble}
\makeatother

\begin{abstract}

In this work we explore the intersection fairness and robustness in the context of ranking: \textit{when a ranking model has been calibrated to achieve some definition of fairness, is it possible for an external adversary to make the ranking model behave unfairly without having access to the model or training data?} To investigate this question, we present a case study in which we develop and then attack a state-of-the-art, fairness-aware image search engine using images that have been maliciously modified using a \textit{Generative Adversarial Perturbation} (GAP) model~\cite{poursaeed2018generative}.
%Our image search engine uses a state-of-the-art MultiModal Transformer (MMT)~\cite{geiglemmtretreival} retrieval model and a fair re-ranking algorithm (FMMR \cite{karako2018using}) that aims to achieve demographic group fairness.
%We then train a generative adversarial perturbation (GAP) model to strategically insert human-imperceptible perturbations into images.
These perturbations attempt to cause the fair re-ranking algorithm to unfairly boost the rank of images containing people from an adversary-selected subpopulation.

We present results from extensive experiments demonstrating that our attacks can successfully confer significant unfair advantage to people from the majority class relative to fairly-ranked baseline search results. We demonstrate that our attacks are robust across a number of variables, that they have close to zero impact on the relevance of search results, and that they succeed under a strict threat model. Our findings highlight the danger of deploying fair machine learning algorithms in-the-wild when (1) the data necessary to achieve fairness may be adversarially manipulated, and (2) the models themselves are not robust against attacks.

\end{abstract}

%\settopmatter{printfolios=true}

\maketitle

\section{Introduction}

The machine learning (ML) community has awoken to concerns about the \textit{fairness} of ML models, \ie the elimination of unjustified bias against specific groups of people from models. There is now extensive literature documenting unfairness in deployed ML systems~\cite{barocas2016big,angwin2019machine,buolamwini2018gender} as well as techniques for training fair classification~\cite{kamishima2012fairness,menon2018cost,goel2018non,huang2019stable} and ranking~\cite{zehlike2017fa,celis2018ranking,singh2018fairness} models. Companies are adopting and deploying fair ML systems in many real-world contexts~\cite{basu-2020-lighthouse,fairnessflow,wilson2021building}.

Most ML systems that strive to achieve demographic fairness are dependent on high-quality demographic data to control for unjustified biases~\cite{bellamy2019ai,vasudevan2020lift}. Recent work has highlighted how critical this dependency is by showing how \textit{unintentional errors} in demographic data can dramatically undermine the objectives of fair ranking algorithms~\cite{ghosh2021fair}. 

Another serious concern in the ML community is model \textit{robustness}, especially in the face of clever and dedicated adversaries. The field of adversarial ML has demonstrated that seemingly accurate models are brittle when presented with maliciously crafted inputs~\cite{szegedy2013intriguing, carlini2017towards}, and that these attacks impact models across a variety of contexts~\cite{szegedy2013intriguing, he2018adversarial, dai2018adversarial, behjati2019universal}. The existence of adversarial ML challenges the use of models in real-world deployments.

In this work we explore the intersection of these two concerns---fairness and robustness---in the context of ranking: \textit{when a ranking model has been carefully calibrated to achieve some definition of fairness, is it possible for an external adversary to make the ranking model behave unfairly without having access to the model or training data?} In other words, can attackers \textit{intentionally} weaponize demographic markers in data to subvert fairness guarantees? 

% Fair ranking models that require knowledge of group membership are vulnerable to poor inference. In the absence of rich demographic labels, companies resort to commercially available demographic inference models to obtain demographic labels which are then used for such fair ranking. Recent work shows that because such commercially available models do not generalize well and have poor accuracies, it has a knock-on effect on the fair ranking models too, causing them to behave unfairly towards demographic minorities \cite{ghosh2021fair}. The intersection of adversarial ML and algorithmic fairness has started gaining some research interest as well. For example, recent work by \citet{zhou2020adversarial} hypothesized that targeted attacks on selected items in a ranked list might be possible using universal adversarial perturbations. 

To investigate this question, we present a case study in which we develop and then attack a fairness-aware image search engine using images that have been maliciously modified with \textit{adversarial perturbations}. We chose this case study because image retrieval based on text queries is a popular, real-world use case for neural models (\eg Google Image Search, iStock, Getty Images, etc.), and because prior work has shown that these models can potentially be fooled using adversarial perturbations~\cite{zhou2020adversarial} (although not in the context of fairness). To strengthen our case study, we adopt a strict threat model under which the adversary cannot \textit{poison} training data~\cite{jagielski2020subpopulation} for the ranking model, and has no knowledge of the ranking model or fairness algorithm used by the victim search engine. Instead, the adversary can only add images into the victim's database \textit {after} the image retrieval model is trained.

\begin{figure*}[t]
    \centering
    \includegraphics[width=0.9\textwidth,keepaspectratio]{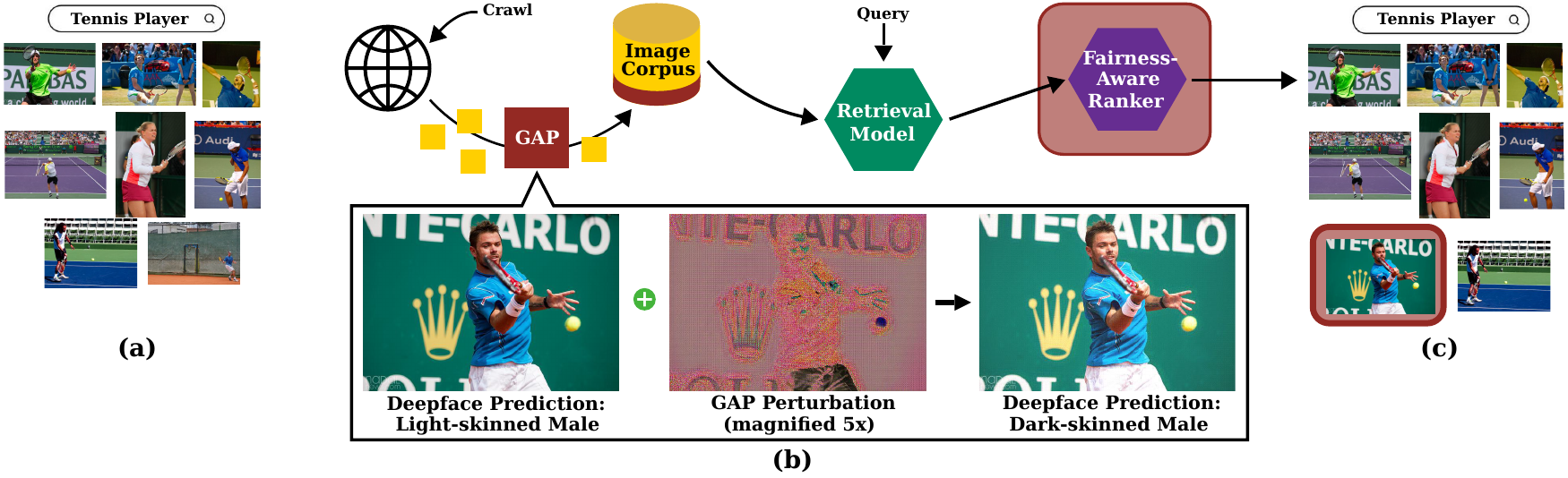}
    \Description{A diagram showing our attack approach. Subfigure \textbf{(a)} shows example search results from an image search engine for the query ``tennis player''. The example results are a grid of eight photographs of people playing tennies. This search engine attempts to provide demographically-fair results, and at this point no images in the corpus have been adversarially perturbed. Subfigure \textbf{(b)} depicts a flowchart of the search engine crawling and indexes new images from the web. As it does so it collects images that have been adversarially perturbed using a GAP model. We show a real example of one image (a male tennis player) before and after applying the generated perturbation, which causes the Deepface model~\cite{taigman2014deepface} to misclassify this person's skin tone. The perturbation itself looks like semi-random noise, and the resulting noised image is visually indistinguishable from the original unperturbed image. Subfigure \textbf{(c)} shows the response to a future query for ``tennis player'' from the image search engine. In this case, the retrieval model will identify relevant images, some of which are perturbed. The fairness-aware ranker (the target of the attack, highlighted in red in the flowchart) mistakenly elevates the rank of an image containing a light-skinned male (also highlighted in red) because it misclassifies them as dark-skinned due to the perturbations.}
    \caption{A diagram showing our attack approach. \textbf{(a)} shows example search results from an image search engine for the query ``tennis player''. This search engine attempts to provide demographically-fair results, and at this point no images in the corpus have been adversarially perturbed. \textbf{(b)} as this search engine crawls and indexes new images from the web, it collects images that have been adversarially perturbed using a GAP model. We show a real example of one image before and after applying the generated perturbation, which causes the Deepface model~\cite{taigman2014deepface} to misclassify this person's skin tone. \textbf{(c)} in response to a future query for ``tennis player'', the retrieval model will identify relevant images, some of which are perturbed. The fairness-aware ranker (the target of the attack, highlighted in red) mistakenly elevates the rank of an image containing a light-skinned male (also highlighted in red) because it misclassifies them as dark-skinned due to the perturbations.} 
    %schematic showing sample results from our fairness-aware search engine with and without our unfairness inducing attack. \textbf{(a)} shows the straightforward case where unperturbed images are crawled from the web and inserted into the retrieval corpus. In \textbf{(b)}, a portion of the images are perturbed by a Generative Adversarial Perturbation (GAP) model before being added to the retrieval corpus. This causes more light-skinned men to appear in the top eight search results, making them less demographically diverse than the results in \textbf{(a)}. \textbf{(c)} shows an instance of GAP induced misclassification by the Deepface model~\cite{taigman2014deepface}. The generated perturbation includes clusters around the face and exposed skin areas.}
    \label{fig:gapdiagram}
    %\vspace*{-9pt}
\end{figure*}

For our experiments, we develop an image search engine that uses a state-of-the-art MultiModal Transformer (MMT)~\cite{geiglemmtretreival} retrieval model and a fair re-ranking algorithm (FMMR~\cite{karako2018using}) that aims to achieve demographic group fairness on the ranked list of image query results without ever explicitly using demographic labels. Under normal circumstances, where the images are unperturbed, our search engine returns demographically balanced sets of images in response to free text queries. We then train a Generative Adversarial Perturbation (GAP) model~\cite{poursaeed2018generative} that learns from pretrained demographic classifiers to strategically insert human-imperceptible perturbations into images. These perturbations attempt to cause FMMR to unfairly boost the rank of images containing people from an adversary-selected subpopulation (\eg light-skinned men). \autoref{fig:gapdiagram} shows example image search results produced by our search engine in response to the query ``Tennis Player'', with and without our attack. 

We present results from extensive experiments demonstrating that our attacks can successfully confer significant unfair advantage to people from the majority class (light-skinned men, in our case)---in terms of their overall representation and position in search results---relative to fairly-ranked baseline search results. We demonstrate that our attack is robust across a number of variables, including the length of search result lists, the fraction of images that the adversary is able to perturb, the fairness algorithm used by the search engine, the image embedding algorithm used by the search engine, the demographic inference algorithm used to train the GAP models, and the training objective of the GAP models. Additionally, our attacks are \textit{stealthy}, \ie they have close to zero impact on the relevance of search results.

In summary, we show that GAPs can be used to subvert fairness guarantees in the context of fair image retrieval. Further, our attack is successful under a highly restricted threat model, which suggests that more powerful adversaries will also be able to implement successful attacks. We hypothesize that similar attacks may be possible against other classes of ML-based systems that (1) rely on highly parameterized models and (2) make fairness decisions for inputs that are based on data controlled by adversaries.

The goal of our work is not to hinder or deter the adoption of fair ML techniques---we argue that fair ML techniques must be adopted in practice. Rather, our goal is to demonstrate that fairness guarantees can potentially be weaponized so that the research community will be energized to develop mitigations, \eg by making models more robust, and by adopting high-quality sources of demographic data that are resistant to manipulation. To facilitate mitigation development without arming attackers, we plan to release our code and data to researchers on request.

% \begin{figure*}[t]
% \centering
% \begin{subfigure}[b]{0.30\textwidth}
%     \includegraphics[width=\linewidth]{Figures/perturbation.png}
%     \caption{Computed Perturbation.}
%      \label{fig:perturb}
%   \end{subfigure}
%   \quad
%   \begin{subfigure}[b]{0.30\textwidth}
%     \includegraphics[width=\linewidth]{Figures/originalImages.png}
%     \caption{Images without perturbation.}
%   \end{subfigure}
%   \quad
%   \begin{subfigure}[b]{0.30\textwidth}
%     \includegraphics[width=\linewidth]{Figures/perturbedImages.png}
%     \caption{Perturbed images.}
%   \end{subfigure}
%   \caption{Comparison of unperturbed and perturbed images.}
%   \label{fig:compPerturb}
% \end{figure*}

%---------------------------------------------------------------------------------------------
%---------------------------------------------------------------------------------------------
%---------------------------------------------------------------------------------------------
%---------------------------------------------------------------------------------------------

\section{Background}

\subsection{Fairness in Ranking}

Algorithmic decision making is permeating modern life, including high-stakes decisions like credit lending \citep{beck2018sex}, bail granting \citep{angwin2019machine}, hiring \citep{wilson2021building}, etc. While these systems are great at scaling up processes with human bottlenecks, they also have the unintended property of embedding and entrenching unfair social biases (\eg sexism and homophobia).
%like racism, sexism, homophobia, ableism, ageism, or religious intolerance.
In response, there is a growing body of academic work on ways to detect algorithmic bias \citep{barocas2016big,ghosh2021characterizing} and develop classes of fair algorithms, for instance classification~\cite{kamishima2012fairness,menon2018cost,goel2018non,huang2019stable}, causal inference~\cite{loftus2018causal,nabi2018fair}, word embeddings~\cite{bolukbasi2016man,brunet2019understanding}, regression~\cite{agarwal2019fair,berk2017convex}, and retrieval/ranking~\cite{zehlike2017fa,celis2018ranking,singh2018fairness}. There is also a growing body of legal work and legislative action around the globe \cite{usreg,eureg,canadareg,ukreg,nzreg} to tackle algorithmic bias.

% \cwnote{This is going to FAccT, so we need more here. Four citations for detecting bias and fair algorithms isn't going to cut it.}

% \subsection{}

In this study we focus on fair Information Retrieval (IR) algorithms---a class of algorithms that have received comparatively less attention than classification algorithms in the literature. Initial studies that examined fair IR proposed to solve this in a binary context, \ie make a ranked list fair between two groups~\cite{celis2018ranking,zehlike2017fa}. Subsequent work uses constrained learning to solve ranking problems using classic optimization methods~\cite{singh2018fairness}. There are also methods that use pairwise comparisons~\cite{googlepairwisefair} and describe methods to achieve fairness in learning-to-rank contexts~\cite{morik2020controlling,zehlike2020reducing}.

In industrial settings, researchers at LinkedIn have proposed an algorithm that uses re-ranking in post-processing to achieve representational parity~\cite{geyik2019fairness}. However, recent work by~\citet{ghosh2021fair} shows how uncertainty due to incorrect inference of protected demographic attributes can undermine fairness guarantees in IR contexts. Fairness methods that do not require explicit demographic labels at runtime are, as of this writing, an emerging area of focus in classification~\cite{lahoti2020fairness} and ranking~\cite{karako2018using,romanov2019s}. One example that has been studied at large-scale is Shopify's Fair Maximal Marginal Relevance (FMMR) algorithm~\cite{karako2018using}, which we describe in more detail in \autoref{sec:fmmr}. In this study, we examine the robustness of Shopify and LinkedIn's fair ranking algorithms.

%\matthew{can we justify this decision a bit more, e.g. largest scale adoption or something}

% \cwnote{Need a sentence or two here talking about Shopify, drawing parallels to other fair-without-demographics work, and then pointing to the detailed description later in the paper.}

\subsection{Adversarial Machine Learning}
\label{sec:back:adversarialML}

Adversarial ML is a growing field of research that aims to develop methods and tools that can subvert the objectives of ML algorithms. For example, prior research has highlighted that deep learning models are often not robust when presented with inputs that have been intentionally, maliciously crafted~\citep{goodfellow2014explaining, moosavi2016deepfool, papernot2016technical, carlini2017towards, vorobeychik2018adversarial, wu2020making,shan2020fawkes}.

Several proposed defenses against state-of-the-art adversarial ML attacks have been defeated~\citep{athalye2018obfuscated, tramer2020adaptive}, and \textit{adversarial examples} (\ie maliciously crafted inputs) have been shown to transfer across models performing similar tasks~\citep{liu2016delving, transferable}. The most promising defense method, adversarial training, is computationally expensive and imperfect---it results in decreased standard accuracy while still having a somewhat low adversarial accuracy~\citep{ganin2016domain, kurakin2016adversarial, shafahi2019adversarial}. As such, adversarial ML presents a significant hurdle to deploying neural models in sensitive real-world systems.

Our work considers adversarial ML attacks on IR systems. Previous work has demonstrated successful attacks on image-to-image search systems~\cite{zhou2020adversarial, liu2019afraid}, allowing an adversary to control the results of targeted image queries using localized patches~\cite{brown2017adversarial} or \textit{universal adversarial perturbations\footnote{A Universal Adversarial Perturbation (UAP) is a an adversarial perturbation that generalizes to all classes of a target classifier, \ie one patch to untargeted attack as many classes as possible.}}~\cite{li2019universal}. Other work has demonstrated attacks on text-to-text retrieval systems~\cite{raval2020one} and personalized ranking systems~\cite{he2018adversarial}. Work by \citet{zhou2020adversarial} hypothesized that targeted attacks on selected items in a ranked list might be possible using universal adversarial perturbations. None of these works consider compromising text-to-image models or group fairness objectives, as we do in this study.

Prior work has demonstrated adversarial ML attacks against fairness objectives of ML systems at \textit{training time}. In these attacks, the adversary supplies \textit{poisoned} training data, which then results in models that either compromise the accuracy of targeted subgroups~\cite{jagielski2020subpopulation, mehrabi2020exacerbating} or exacerbate pre-existing unfairness between subgroups~\cite{chang2020adversarial, solans2020poisoning}. Specific to classification, there exists theoretical work that shows how to learn fair models when sensitive attributes are noisy~\cite{celis2021fair} or corrupted in a poisoning attack~\cite{celis2021fair1}, but they do not consider ranking.
%(despite specific training objectives to ensure fairness across groups)

Adversarial ML attacks at \textit{test time}---\ie after training a model using non-malicious data---that we consider in this work are relatively unexplored in fairness settings. \citet{nanda2021fairness} show that adversarial ML attacks can harm certain subpopulations more than others in classification tasks. However, while this is an important observation, the harms suggested by this work may be difficult to realize in practice, as they only involve disparity between examples that are adversarially corrupted. By contrast, our work shows that test-time attacks can harm fairness for benign data when launched in a ranking setting.

%\subsection{Research Gaps and Motivation}

%The intersection between ML fairness and adversarial ML is where our proposed work is situated. While fairness interventions and fair ML algorithms exist aplenty \citep{hardt2016equality,zafar2017fairness,berk2017convex,celis2019classification}, work is only beginning to be published about the robustness of these models \citep{nanda2021fairness}. In light of the tension between fairness and robustness of models in the presence of an adversary \citep{chang2020adversarial}, we pose the following question: \textbf{\itshape When a model has been carefully calibrated to achieve some definition of fairness, is it possible for a malicious actor to attack the trained model to make the model behave unfairly}? In contrast to previous work which tests the unfairness of models due to data poisoning \citep{jagielski2020subpopulation, solans2020poisoning, chang2020adversarial}, we aim to investigate attacks by external parties in a black box scenario, thereby making the threat model more applicable in practice. 

%---------------------------------------------------------------------------------------------
%---------------------------------------------------------------------------------------------
%---------------------------------------------------------------------------------------------
%---------------------------------------------------------------------------------------------

\section{Methodology}
\label{sec:method}

We now present the plan for our study. First, we introduce our application context and the threat model under which an attacker will attempt to compromise the application. Second, we discuss the IR models and algorithms underlying our fairness-aware image search engine. Third, we discuss our strategy for attacking this search engine using GAPs.

\subsection{Context and Threat Model}
\label{sec:meth:threatmodel}

In this study, we consider the security of a fairness-aware image search engine. This search engine indexes images from around the web (either automatically via a crawler or from user-provided submissions) and provides a free-text interface to query the image database. Examples of image search engines include Google Image Search, iStock and Getty Images, and Giphy. In our case, the image search engine attempts to produce results that are both relevant to a given query and fair, according to some fairness objective. One example fairness objective is demographic representativeness, \ie for search results that contain images of people.

We consider a malicious image curator (\eg Imgur, 4chan, or similar) with a large database of \textit{perturbed} images that are eventually scraped or uploaded into the victim image search engine's index.\footnote{An adversarial image curator is also the threat model assumed for clean label poisoning attacks~\cite{shafahi2018poison, turner2018clean}. This adversarial image curator may perturb copies of images taken from the web or original images that they author. This setup is also used by \cite{shan2020fawkes} as a defensive method against unauthorized models.} Our adversarial image curator's goal is to perturb the images in their database to subvert the fairness guarantees of the downstream retrieval system. We assume that the adversary does not have any knowledge of the internals of the ranking system (\eg what retrieval model is used, other images in the index, or which fairness algorithm is used).

This threat model constitutes a strict, but realistic, limitation on our adversary. Notice that this threat model would also apply if the image search engine was compromised, giving the adversary access to underlying models and the entire dataset of images. We consider both adversaries in our experiments. We also note that, if the adversary only seeks to target a small set of queries, they need only control a fraction of the images matching each query, rather than a fraction of the entire image database. This is useful for the adversary in the case that not all queries are equally sensitive. 

%---for our study, we focus on the Fair-MMR algorithm, as prior work has demonstrated the vulnerability of demographic-aware algorithms.\matthew{this used to not be an assumption but now it kind of is because we only use Shopify?}.

\subsection{Building an Image Search Engine}

We now turn our attention to building a realistic image search engine that will serve as the victim for our attacks. 
%In the following sections we introduce the image retrieval model (MultiModal Transformer~\cite{geiglemmtretreival}) and fairness algorithm (Fair Maximal Marginal Relevance~\cite{karako2018using}) that we adopt.

\subsubsection{Image Retrieval from Text Queries}
\label{sec:retrieval}

The first choice we make for this study is to select an image retrieval model. There are several frameworks for image retrieval in the literature, starting from tag-based matching \cite{lerman2007personalizing} to state-of-the-art vision-language transformers \cite{lu2019vilbert,li2020oscar}. For the purpose of this paper, we used a MultiModal Transformer (MMT)~\cite{geiglemmtretreival} based text-image retrieval model. This model consists of two components: a fast (although somewhat lower quality) retrieval step that identifies a large set of relevant images, followed by a re-ranking step that selects the best images from the retrieved set. Concretely, the user provides a string $q$ that queries into a database $D$ of $n$ images. For the retrieval step, the query string is encoded with an embedding function $f_q$ to produce an embedding $v_q$, and all images in $D$ have pre-computed embeddings from an embedding function $f_I$. The cosine distance between $v_q$ and all embeddings of $D$ are computed to collect some large set $D_q$ of size $n' \le n$ plausible image matches. These images are then ranked according to a joint model $f_j$ that takes both the query and an image as input, returning scores $\lbrace s_i \rbrace_{i=1}^{n'}$ indicating how well each image $D_q[i]$ matches the query. These scores are used to produce the final ranking.

Note that the MMT model is not designed to be ``fair'' in any normative sense. To achieve fairness, results from the model must be re-ranked, which we describe in the next section. Thus, the MMT model is not the target of our attacks, since it is not responsible for implementing any fairness objectives.

\subsubsection{Fairness-aware Re-ranking}
\label{sec:fmmr}

The second choice we make for this study is selecting an algorithm that takes the output of the image retrieval model as input and produces a fair re-ranking of the items. In fairness-aware re-ranking, a ranking function $f_r(D, q)$ is post-processed to achieve fairness according to some subgroup labels on the dataset $D=\lbrace s_i, x_i\rbrace_{i=1}^n$, where $s_i$ denotes the score of the $i^{\textrm{th}}$ item (the heuristic score according to which the list is sorted) and $x_i$ denotes the item to be ranked.

%In this study, we use demographic characteristics as the subgroup labels for fair re-ranking. Additionally, we use intersectional attributes for $a_i$ obtained via Cartesian products of marginal attributes, like in~\cite{ghosh2021characterizing}.

%\paragraph{Without Using Demographic Labels}

The re-ranking algorithm we adopt is Fair Maximal Marginal Relevance (FMMR)~\cite{karako2018using}, which was developed and used at Shopify for representative ranking of images. FMMR builds on the Maximal Marginal Relevance~\cite{carbonell1998use} technique in IR that seeks to maximize the information in a ranked list by choosing the next retrieved item in the list to be as dissimilar to the current items present in the list as possible. MMR introduces a hyperparameter that allows the operator to choose the trade-off between similarity and relevance.

FMMR modifies the ``similarity'' heuristic from MMR to encode for similarity in terms of demographics, with the idea being that the next relevant item chosen to be placed in the re-ranked list will be as demographically different from the existing images as possible. Similarity is calculated using image embeddings, for which we examine three models: Faster R-CNN~\cite{ren2015faster}, InceptionV3~\cite{szegedy2015going}, and ResNet18~\cite{he2016deep}. We fix the trade-off parameter $\lambda$ at 0.14 as that is the value used by \citet{karako2018using} in their FMMR paper.

It is notable that FMMR does not require demographic labels of people in images to perform fair re-ranking, since it uses a heuristic that only relies on embeddings.  Indeed, FMMR comes from a class of fair ranking algorithms that all use the inherent latent representations of the objects for their re-ranking strategy~\cite{romanov2019s,karako2018using}. That said, since FMMR attempts to maximize the distance from the centroids of the embeddings of different demographic groups, it can be thought of as performing indirect demographic inference on individuals in images.

Additionally, we also evaluated our attacks against a second fairness-aware re-ranking algorithm, DetConstSort~\cite{geyik2019fairness}, developed by and deployed at LinkedIn in their talent search system. Unlike FMMR, DetConstSort explicitly requires demographic labels for the items it is trying to fairly re-rank.
%As a result, we might expect our attack, which targets demographic inference models, to perform well against DetConstSort.
However, prior work~\cite{ghosh2021fair} shows that DetConstSort has significant limitations when demographic inference is used rather than ground-truth demographic labels, making it unfair even without perturbed images. As a result, evaluating an attack against DetConstSort is not meaningful, and we defer our discussion of DetConstSort to \autoref{sec:detconstsort}.

\subsection{Attack Construction}
\label{sec:meth:attackconstruction}

Having described our search engine, we are ready to turn our attention to our attack. First, we introduce the demographic inference models (Deepface~\cite{taigman2014deepface} and  FairFace~\cite{karkkainen2019fairface}) that we use to train our attack. Next, we describe how we generate adversarial perturbations from a demographic inference model, modifying images in a way that is imperceptible to human eyes, yet significant enough to fool the fair re-ranking algorithm of our search engine.

\subsubsection{Demographic Inference Algorithms}
\label{sec:DI}

For large-scale datasets such as images scraped from the web, demographic meta-data for people in the images is (1) not readily available and (2) prohibitively expensive to collect through manual annotation~\cite{andrus2021measure, bogen2020awareness}.
%Thus, to make our image search engine realistic, we use a demographic inference model to generate protected attribute labels that are used for the purpose of fair re-ranking. 
Pipelines using demographic inference are commonly used in practice when demographic labels are not available. For example, the Bayesian Improved Surname Geocoding (BISG) tool is used to measure fairness violations in lending decisions~\cite{BISG, BISG-use1}, and it relies on inferred demographic information. This makes attacks on demographic inference models a natural candidate for adversely affecting ranking fairness.

We consider two image demographic inference models to train our attacks:
\begin{enumerate}
    \item Deepface~\cite{taigman2014deepface} is a face recognition model for gender and race inference developed by Facebook. We use its public wrapper~\cite{serengil2020lightface}, which includes models fine tuned on roughly 22,000 samples for race and gender classification.
    
    \item FairFace~\cite{karkkainen2019fairface} is a model designed for race and gender inference, trained on a diverse set of 108,000 images.
\end{enumerate}

Since both of these models infer race/ethnicity, we used a mapping to infer skin tone, since we could not find commercially available algorithms to infer skin tones from human images.\footnote{The mapping we used is: White, East Asian, Middle Eastern → Light, and Black, South Asian, Hispanic → Dark. We acknowledge that this is a crude mapping, but it enabled us to train a successful attack.} We also use these models to infer demographics as input to the DetConstSort algorithm, matching the pipeline of \cite{ghosh2021fair}, which we discuss in \autoref{sec:detconstsort}.

\subsubsection{Subpopulation Generative Adversarial Perturbations}
\label{sec:cgap}

Recall our adversarial image curator's goal: to produce a database of malicious images that, when indexed by our image search engine, undermine its purported fairness guarantees. Concretely, this means fooling the fair re-ranker such that it believes a given set of search results is fair across two or more subgroups, when in fact the results are unfair because some subgroups are under- or over-represented. Additionally, these malicious images must (1) retain their relevance to a given query and (2) not be perceived as ``manipulated'' to human users of the search engine.

Prior work (see \autoref{sec:back:adversarialML}) has demonstrated that neural image classification models can be fooled by adding \textit{adversarial perturbations} to images. At a high-level, the adversary's goal is to train a model that can add noise to images such that specific latent characteristics of the images are altered. In our case, these altered characteristics should impact the image embeddings calculated by the image embedding model (\eg InceptionV3) that FMMR relies upon to do fair re-ranking.

Running an adversarial perturbation algorithm on each of the images in the adversary's database would be prohibitive, as these algorithms involve computationally expensive optimization algorithms that are not practical at the scale of an entire database. We avoid this limitation by training a Generative Adversarial Perturbation (GAP) model~\cite{poursaeed2018generative}. A GAP model $f_{\text{GAP}}$ takes a clean image as input and returns a perturbed image that is misclassified by some target model $f_{\text{targ}}$. This replaces the per-image optimization problem with a much less expensive forward pass of $f_{\text{GAP}}$. Training the GAP is a one time expense for the adversary, amortized over the large number of image perturbations done later. Universal Adversarial Perturbations (UAPs)~\cite{moosavidezfooli2017universal} are another approach to amortizing runtime, but require all images to be the same dimensions---an unrealistic assumption for real-world image databases.\footnote{A UAP can also be seen as a GAP, where $f_{\text{GAP}}(x)=x+\delta$ for a fixed $\delta$. Therefore, we expect that a GAP will perform strictly better than a UAP.}

Having motivated the choice of a GAP model for our attack, we now consider the problem of impacting fairness by attacking the fair re-ranking algorithm used by a victim search engine. We choose to design a GAP to target a demographic inference model $f_{\text{DI}}$.\footnote{Recall that, per our threat model in \autoref{sec:meth:threatmodel}, the attacker does not know what fair re-ranking algorithm is used by the victim and thus cannot train against it directly.} This will produce perturbations that, to a deep image model, make an image of a person from one demographic group appear to be from a different demographic group. This attack would heavily impact a demographic-aware re-ranking algorithm such as DetConstSort~\cite{geyik2019fairness} (see \autoref{sec:detconstsort}) if it used an accurate demographic inference algorithm to produce annotations.

Although FMMR does not use annotations, we show in \autoref{sec:results} that our attack is still successful at compromising FMMR's fairness guarantees. Our attack can be seen as an application of the \textit{transferability property} of adversarial examples. Additionally, training our GAP against a demographic inference model causes our attack to be independent of the ranking algorithm and image corpus used by the victim search engine, both of which are strong adversarial assumptions. 

%It also leverages the observation made by existing work that inaccurate demographic inference alone is sufficient to compromise the performance of fair ranking systems~\cite{ghosh2021fair}.

In designing our GAP to compromise fairness, we first note that an attack that simply forces a $f_{\text{DI}}$ to make arbitrarily many errors may not impact fairness. For example, suppose the image database contained two subpopulations, the advantaged class $A$ and the disadvantaged class $B$. Suppose the attack causes $f_{\text{DI}}$ to misclassify all members of $B$ as $A$ and all members of $A$ as $B$. This is the best possible result of an attack on the demographic inference algorithm, but results in no changes to a fair ranking algorithm---it will simply consider $A$ to be the disadvantaged class, and thus produce the same ranking! For this reason, our adversary must incorporate subpopulations into the attack. To do so, we propose the Class-Targeted Generative Adversarial Perturbation (CGAP):

\begin{definition}[CGAP]
We consider a loss function $\ell$, target model $f_{\text{targ}}$, distribution $\mathcal{D}$ over inputs $x$ and outputs $y$. The adversary provides a source class $y_s$ and target class $y_t$. Then the CGAP model $f_{\text{CGAP}}$ is a model that takes as input an image $x$ and returns an image $x'$, minimizing the following loss functions:
\begin{align*}
\ell_{\text{CGAP}}^{s}(\mathcal{D}) = \mathbb{E}_{(x, y) \sim \mathcal{D}}[\ell(f_{\text{CGAP}}(x), y_t; f_{\text{targ}}) | y=y_s], \\
\ell_{\text{CGAP}}^{r}(\mathcal{D}) = \mathbb{E}_{(x, y) \sim \mathcal{D}}[\ell(f_{\text{CGAP}}(x), y; f_{\text{targ}}) | y\neq y_s].
\end{align*}
That is, the CGAP should force the demographic inference model to misclassify samples of class $y_s$ to class $y_t$, while maintaining its performance for samples not from class $y_s$.
\end{definition}

We also consider two extensions of this definition. First, we permit the adversary to target multiple classes at once. In the extreme, an adversary may want all samples to be classified to the same class (this approach is proposed by \cite{poursaeed2018generative}). For a demographic inference algorithm, all samples having the same demographic label will cause the fair re-ranking system to have similar performance to an unfair ranking system, as all points will appear to fall into the same subpopulation. The second extension is the untargeted attack, where the CGAP simply increases loss for points from class $y_s$, inducing arbitrary misclassifications. Simultaneously making both relaxations recovers the original untargeted GAP approach. We experiment with both relaxations independently, as well as multiple instantiations of CGAP as defined above.

%---------------------------------------------------------------------------------------------
%---------------------------------------------------------------------------------------------
%---------------------------------------------------------------------------------------------
%---------------------------------------------------------------------------------------------

\section{Experiments}
\label{sec:experiments}

In this section we introduce the dataset we used for our evaluation, describe the setup for our experiments, and define the metrics we use to evaluate our attacks.

\begin{table*}[t]
%\begin{adjustbox}{width=0.9\columnwidth,center}
\resizebox{0.9\textwidth}{!}{%
\centering
\begin{tabular}{@{}llllll@{}}
\toprule
  \textbf{Search Queries} & 
  \textbf{Attack Training} &
  \textbf{Embedding} &
  \textbf{Training Objective} &
  \textbf{Attack Probability} &
  \textbf{Top $k$} \\ 
  \midrule
  \begin{tabular}[c]{@{}l@{}}``Tennis Player''\\ ``Person eating pizza''\\ ``Person at Table''\end{tabular} &
  \begin{tabular}[c]{@{}l@{}}Deepface\\ FairFace\end{tabular} &
  \begin{tabular}[c]{@{}l@{}}F-RCNN\\ InceptionV3\\ ResNet\end{tabular} &
  \begin{tabular}[c]{@{}l@{}}Any→Light Men\\ Light Men→Any\\ Dark Men→Light Men\\ Light Men→Dark Men\end{tabular} &
  0.2, 0.5, 0.7, 1.0 &
  10, 15, 20..., 45, 50 \\
  \bottomrule
\end{tabular}
}
%\end{adjustbox}
\caption{Variables and hyperparameters we used for evaluating our attack.}
\label{tab:experiments}
% %\vspace*{-15pt}
\end{table*}

\subsection{Dataset, Annotation, and Preprocessing}
\label{sec:data}

We use Microsoft's Common Objects in Context (MS-COCO) \cite{lin2014microsoft} as our retrieval dataset, since it contains a variety of images with variable dimensions and depths. This closely mimics what a real-world image search dataset might contain.

To specifically measure for demographic bias, we filter the dataset, keeping only images that contain people. We also need the images to have demographic annotations for fair ranking, so we use an annotated subset of the COCO 2014 dataset, constructed by \citet{zhao2021captionbias}. Similar to prior work~\cite{celis2020implicit, ghosh2021fair}, Zhao~\etal crowdsource skin color (on the Fitzpatrick Skin Type Scale, which the authors simplified to Light and Dark) and binary perceived gender expression for 15,762 images. For the purposes of our experiments we only considered the 8,692 images that contain one person. After filtering, our final dataset consisted of 5,216 Light Men, 2,536 Light Women, 714 Dark Men, and 226 Dark Women.

\subsection{Experimental Setup}
\label{sec:setup}

As a starting point for our experiments, we need to collect ranked lists from our baseline, \textbf{unfair} retrieval system, as described in \autoref{sec:retrieval}. To do so, we run three different search queries on the retrieval system: ``Tennis Player'', ``Person eating Pizza'', and ``Person at table''.  We chose these queries because they all reference a human being, are ethnicity and gender neutral, and are well-supported in the COCO dataset (we picked popular object tags, see \autoref{sec:querychoice}). We set the upper bound in the baseline retrieval system to be 200 images. The three queries return 131, 75, and 124 images, respectively, along with their relevance scores.

\begin{figure*}[t]
    \begin{subfigure}[t]{0.3\textwidth}
        \centering
        \includegraphics[width=\columnwidth,keepaspectratio]{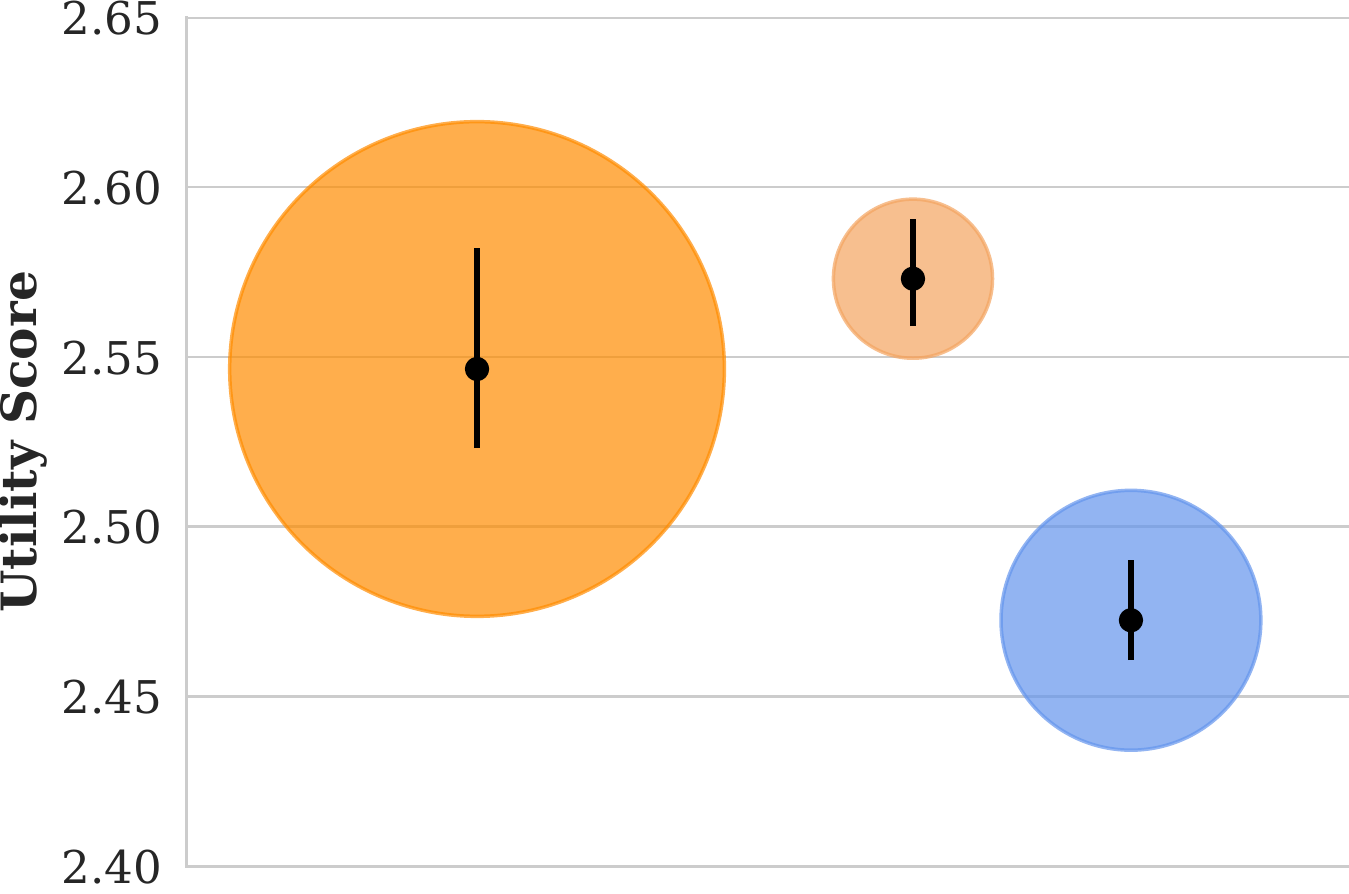}
        \Description{A plot showing the utility scores of images in our corpus matching the query ``tennis player'', along with size estimates of the population broken down by demographic.}
        \caption{Query: ``Tennis Player''}
    \end{subfigure}
    \begin{subfigure}[t]{0.3\textwidth}
        \centering
        \includegraphics[width=\columnwidth,keepaspectratio]{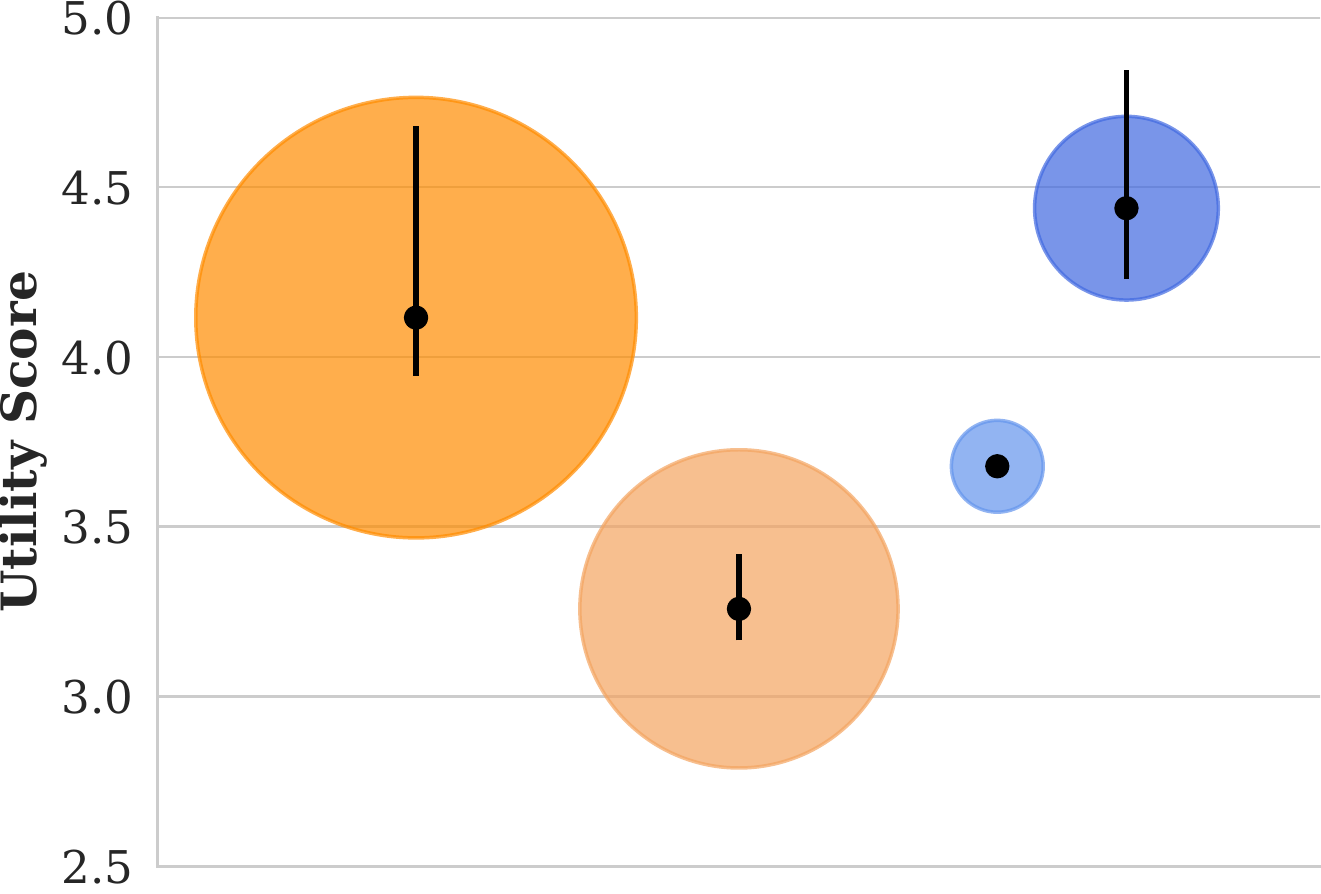}
        \Description{A plot showing the utility scores of images in our corpus matching the query ``person eating pizza'', along with size estimates of the population broken down by demographic.}
        \caption{Query: ``Person eating Pizza''}
    \end{subfigure}
    \begin{subfigure}[t]{0.3\textwidth}
        \centering
        \includegraphics[width=\columnwidth,keepaspectratio]{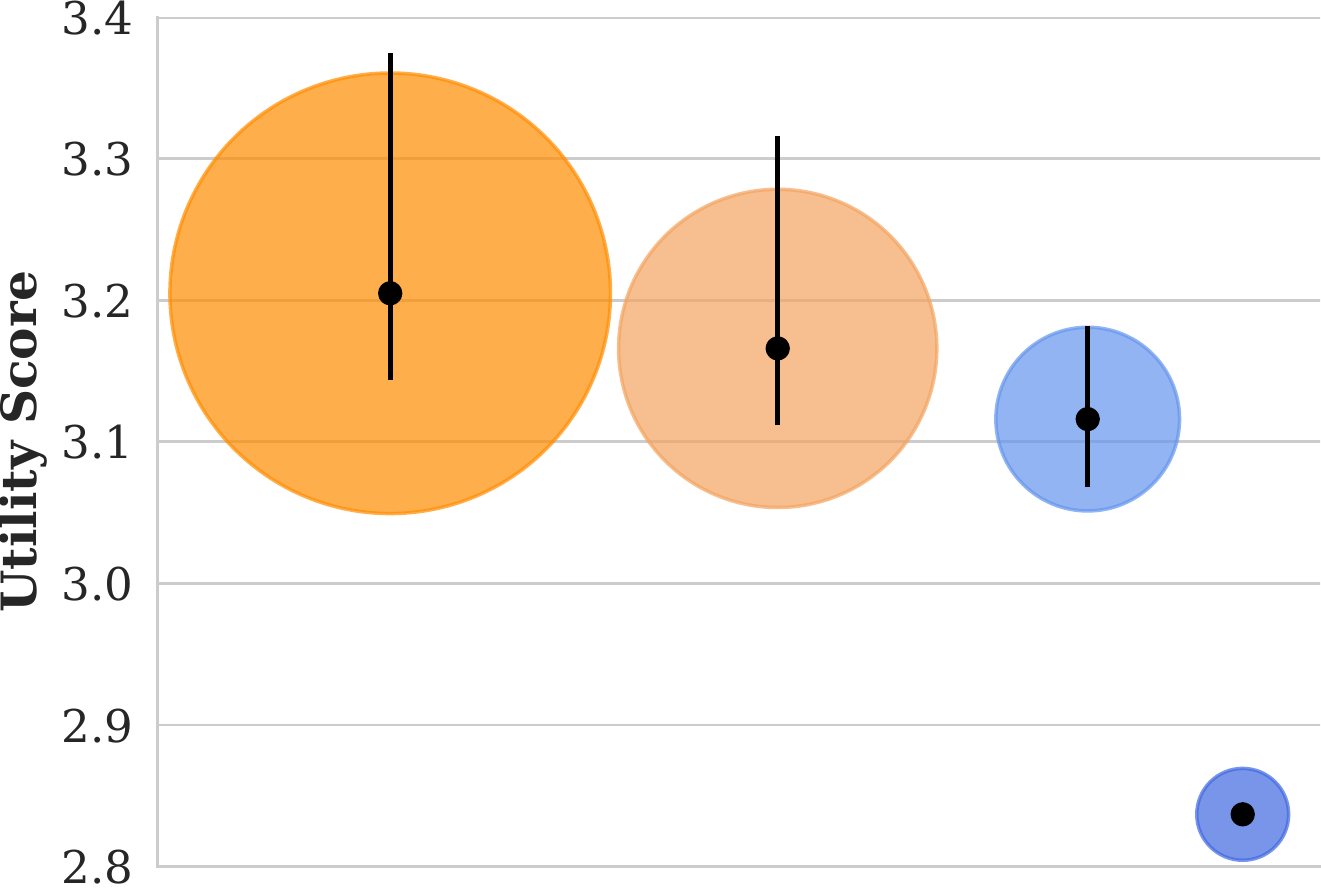}
        \Description{A plot showing the utility scores of images in our corpus matching the query ``person at table'', along with size estimates of the population broken down by demographic.}
        \caption{Query: ``Person at table''}
    \end{subfigure}
    \begin{subfigure}[t]{0.66\textwidth}
    \centering
        \includegraphics[width=\textwidth,keepaspectratio]{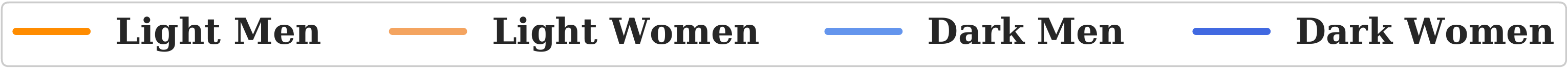}
        \Description{Legend for the utility score figures, color coding demographic four groups: light-skinned men, light-skinned women, dark-skinned men, and dark-skinned women.}
    \end{subfigure}
    \caption{Utility/Relevance score and group size distribution within the top 40 baseline search results for three queries. The black dots represent the average utility score for that group, while the circle size represents the group size. No dark-skinned women appear in the top 40 baseline results for the ``tennis player'' query.\protect\footnotemark}
    \label{fig:circleplot}
    %\vspace*{-9pt}
\end{figure*}

We show the distribution of the relevance scores and the skin color/gender distributions of the images within the top 40 search results for each query in~\autoref{fig:circleplot}. As also shown by~\citet{zhao2021captionbias}, Light Men comprise the overwhelming majority in all three lists, and they also have high relevance scores across the board, meaning that the retrieval system places Light Men near the top of the search results. We call these lists the \textit{baseline} lists.

We also need to produce fair versions of the baseline lists. To do so, we pass the baseline lists for each of our three queries through \textit{FMMR} with the three embedding algorithms, without any adversarial perturbations.
%of the images or using any demographic inference---the demographic attributes for the images in this case were generated by human annotators.
We refer to the nine lists obtained via the fair re-ranker (three queries times three image embedding models) as the \textit{oracle} lists.

\footnotetext{Dark-skinned women do appear in the search results for the query ``female tennis player''. This seems to reflect stereotypical bias~\cite{garg2018word} within the learned-word representations in the MMT model.}

To train our adversarial attacks, we first remove the 330 images in our \textit{oracle} lists from the original dataset, leaving 8,362 images. These 8,362 images were then split randomly into training and testing sets in an 8:2 ratio to train CGAP models for all possible combinations of training objectives and demographic inference algorithms $f_{DI}$ (described in detail below). We ran our experiments on PyTorch with a CUDA backend on two NVIDIA RTX-A6000 GPUs, and trained CGAP models for 10 epochs each, with the $L_{\infty}$ norm\footnote{$L_{\infty}$ is the absolute distance in pixel space any one pixel is changed, i.e. a pixel can at most change by a value of 10 in each color channel.} bound set to 10.

We describe our different training and inference combinations below. \autoref{tab:experiments} shows a summary of the different settings involved during the training and testing of our CGAP attacks.

\subsubsection{Embedding Algorithm}

As we discuss in \autoref{sec:fmmr}, FMMR requires image embeddings. The authors of the original paper used a pretrained InceptionV3 model, which we also adopt. Additionally, we test the performance of FMMR using embeddings generated by pretrained Faster R-CNN and ResNet models. These models are trained for standard image classification tasks and have no inherent concept of demographic groups.

\subsubsection{Attack Training Algorithm}

As detailed in \autoref{sec:meth:attackconstruction}, we train CGAP models to induce adversary-selected misclassifications in two target demographic inference models, denoted as $f_{DI}$: Deepface~\cite{taigman2014deepface} and FairFace~\cite{serengil2020lightface}. These models are trained for demographic inference, and so do not overlap in training objective with the image embedding models for FMMR. The only similarity in architecture between the demographic inference and FMMR embedding models is that FairFace uses a ResNet architecture.

\subsubsection{Training Objectives}

As discussed in \autoref{sec:cgap}, we select certain subpopulations to be systematically misclassified by the two $f_{DI}$ described above. The four CGAPs we train induce misclassifications with the following source-target pairs: Any$\rightarrow$Light Men, where every subgroup was perturbed to be predicted as Light Men; Light Men$\rightarrow$Any, where only Light Men are arbitrarily misclassified; Dark Men$\rightarrow$Light Men, where only Dark Men are misclassified as Light Men; and Light Men$\rightarrow$Dark Men, where only Light Men are misclassified as Dark Men.

\subsubsection{Attack Probability $pr$}

It is a strong assumption that an adversary can perturb the entire image database of a victim search engine. This is only possible if the search engine itself is malicious or it is utterly compromised. Instead, we measure the effect of our attack when the attacker may perturb $pr = $ 20\%, 50\%, 70\% and 100\% of the image database relevant to each query. If a small number of queries are targeted, only few images are required to run the attack.

\subsubsection{Top $k$}

Ranking is very sensitive to position bias~\cite{sapiezynski2019quantifying,ghosh2021fair}, so we measure with different lengths $k$ of the top list, ranging from top 10 to top 50, to gauge our attack's impact on the fair ranking algorithms as final list sizes vary.

\subsection{Evaluation Metrics}

To evaluate the impact of our attacks, we use three metrics that aim to measure (1) representation bias, (2) attention or exposure bias, and (3) loss in ranking utility due to re-ranking. Additionally, we introduce a summarizing meta-metric that enables us to clearly present the impact of our attacks with respect to each metric.

\subsubsection{Skew} The metric we use to measure the bias in representation is called Skew~\cite{geyik2019fairness, ghosh2021fair}. For a ranked list $\tau$, the $\text{Skew}$ for attribute value $a_i$ at position $k$ is defined as:
\begin{equation}
    \text{Skew}_{a_{i}}@k(\tau)= \frac{p_{\tau^{k},a_{i}}}{p_{q,a_{i}}}.
\end{equation}
 $p_{\tau^{k},a_{i}}$ represents the fraction of members having the attribute $a_{i}$ among the top $k$ items in $\tau$, and $p_{q,a_{i}}$ represents the fraction of members from subgroup $a_{i}$ in the overall population $q$. In an ideal, fair representation, the skew value for all subgroups is equal to 1, indicating that their representation among the top $k$ items exactly matches their proportion in the overall population.

\subsubsection{Attention}  Even if all subgroups were fairly represented in the top $k$ ranked items of a list, the relative position of the ranked items adds another dimension of bias---unequal exposure. Previous studies~\cite{nielsen2003usability, mullick2019public} have shown that people's attention rapidly decreases as they scan down a list, with more attention given to the higher ranking items, ultimately dropping to zero attention. 

In this study, we model attention decay using the geometric distribution as done in prior work by \citet{sapiezynski2019quantifying}. We compute attention at the $k^{th}$ rank as:
\begin{equation}
    \text{Attention}_{p}@k(\tau) = 100\times(1-p)^{k-1}\times(p)
\end{equation}
where $p$ is the fraction of total attention given to the top search result. The choice of $p$ is application specific---for this paper we fixed $p$ to be 0.36, based on a study~\cite{googleclicks} that reported that the top result on Google Search receives 36.4\% of the total clicks. We then calculate the average attention per subgroup:
\begin{equation}
    \text{Average attention}_{a_i,\tau} = \frac{1}{|a_{i}|}\sum_{k=1}^{|\tau|}\text{Att(k)} ~\text{where}~ a_k^{\tau} = a_{i}.
\end{equation}
Ideally, in a perfectly fair ranked list, all subgroups should receive equal average attention.

\subsubsection{Normalized Discounted Cumulative Gain.} NDCG is a widely used measure in IR to evaluate the quality of search rankings~\cite{jarvelin2002cumulated}. It is defined as 
\begin{equation}
    \text{NDCG}(\tau)= \frac{1}{Z}\sum_{i=1}^{|\tau|}\frac{s_{i}^{\tau}}{log_{2}(i+1)}
\end{equation}
where $s_{i}^{\tau}$ is the utility score from the MMT retrieval model of the $i^{th}$ element in the ranked list $\tau$ and $Z=\sum_{i=1}^{|\tau|}\frac{1}{log_{2}(i+1)}$. NDCG scores range from 0 to 1, with the latter capturing ideal search results. 

\subsubsection{Summarizing Metric}

For the purpose of quantifying how much unfair advantage our attacks confer on members of the majority class relative to all other classes, we define a new meta-metric called Attack Effectiveness $\eta$. For a given metric $m$ $\in$ \{ Skew, Attention \} and a subgroup $g$, it is defined as:

\begin{equation}
\begin{split}
\eta (m,g)= \text{ \% change in $m$ for subgroup $g$ } - \\ \text{ minimum \% change in $m$ over other subgroups}.
\end{split}
\end{equation}

%\begin{align}
%    \eta (m,g)= & \text{ \% change in $m$ for subgroup $g$ } - \\ & \text{ minimum \% change in $m$ over other subgroups}.\nonumber
%\end{align}

We chose this formulation of $\eta$ for two reasons. First, comparing percentage changes makes the metric scale invariant, which is useful since group sizes vary. Second, comparing to the group that gets the minimum boost ensures that the metric presents the widest fairness disparity, regardless of the total number of groups.

For the purposes of this paper, we set $g$ as Light Men, because they are socially and historically the most advantaged group, and a large $\eta$ for Light Men indicates that the attack causes their ranking to be unfairly boosted relative to the least privileged group. To make sure that the fairness impacts we observe are due to the effectiveness of our attack on the re-ranking algorithms only, the $\eta$ values and the \% change in NDCG are all measured against the \textit{oracle} (\ie fairly re-ranked) lists. Because we compare against the oracle list, all results with attack probability $pr=0$ will have $\eta=0$.

%---------------------------------------------------------------------------------------------
%---------------------------------------------------------------------------------------------
%---------------------------------------------------------------------------------------------
%---------------------------------------------------------------------------------------------

\section{Results}
\label{sec:results}

In this section, we evaluate the impact of our attacks on the fairness guarantees of FMMR. For each set of results we examine how attack effectiveness varies for one particular variable (\eg top $k$, image embedding model, etc.) as the attack probability $pr$ (\ie the fraction of images under adversarial control) varies. When focusing on a particular variable, we present results that are averaged across all other variables and all three of our queries.

\begin{figure*}[t]
    \begin{subfigure}[t]{0.3\textwidth}
        \centering
        \includegraphics[width=\columnwidth,keepaspectratio]{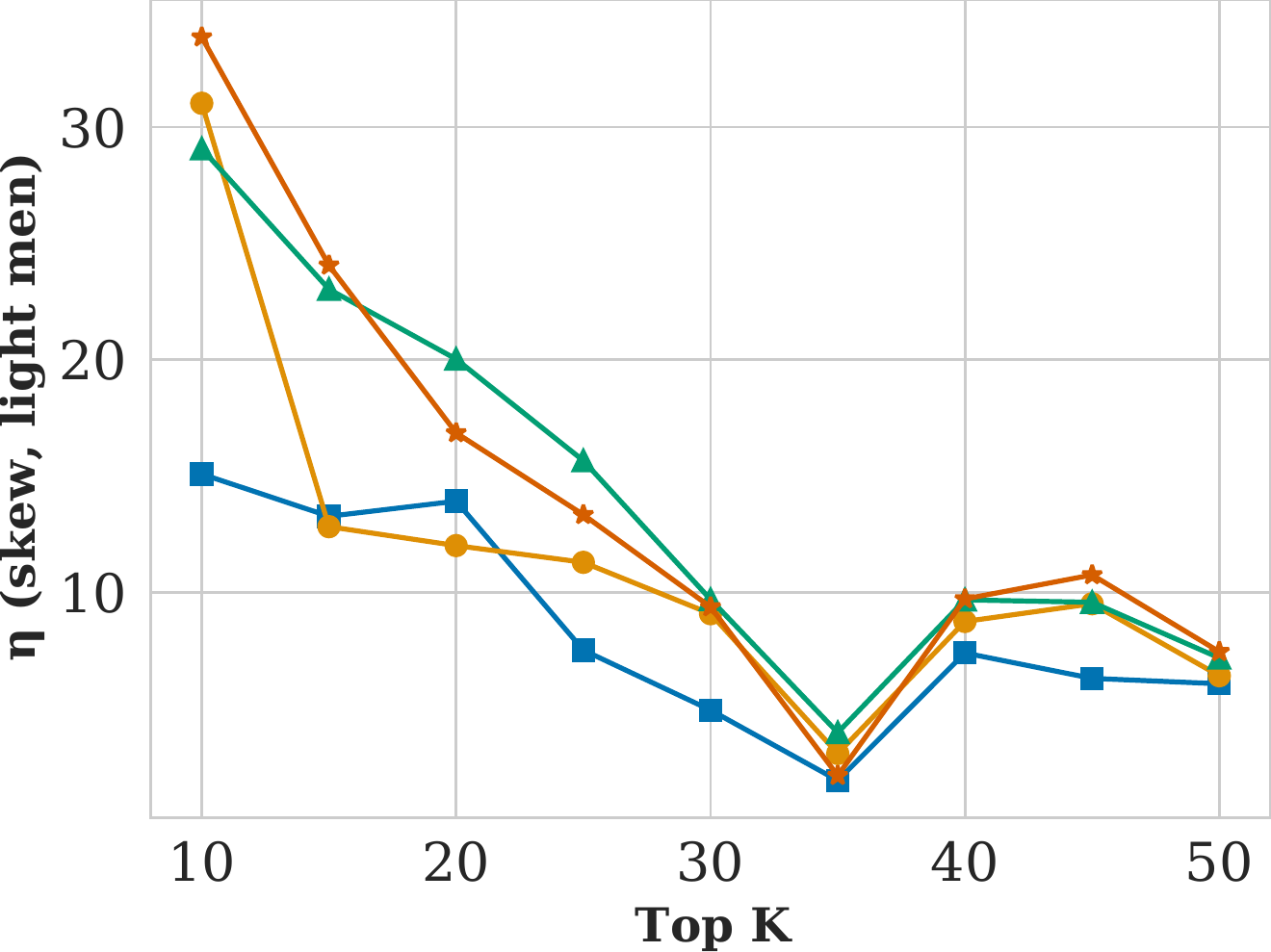}
        \Description{Figure showing how skew varies with respect to search result list length $k$ and attack probability $pr$.}
        \caption{Skew}
    \end{subfigure}
    \begin{subfigure}[t]{0.3\textwidth}
        \centering
        \includegraphics[width=\columnwidth,keepaspectratio]{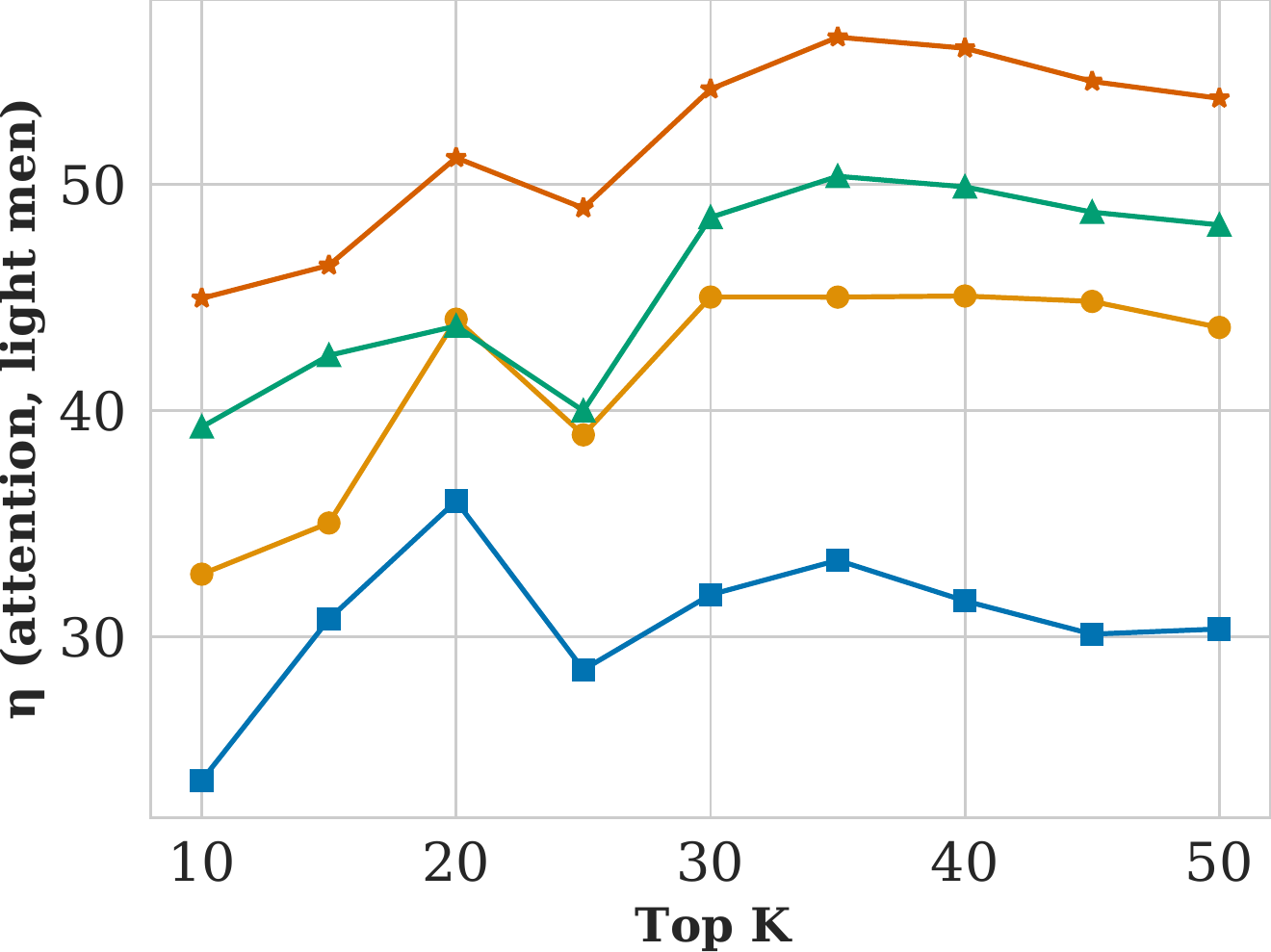}
        \Description{Figure showing how attention varies with respect to search result list length $k$ and attack probability $pr$.}
        \caption{Attention}
    \end{subfigure}
    \begin{subfigure}[t]{0.3\textwidth}
        \centering
        \includegraphics[width=\columnwidth,keepaspectratio]{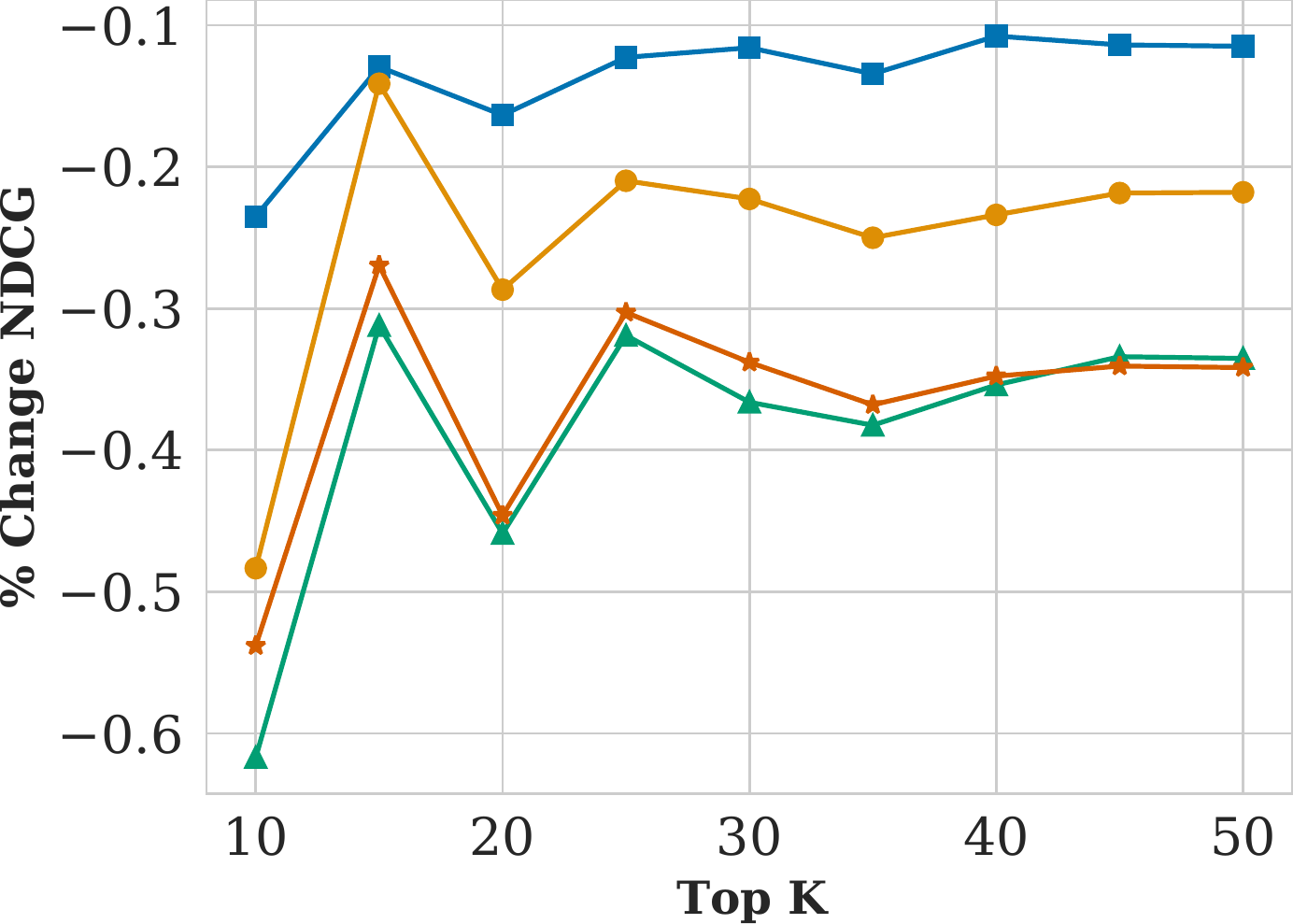}
        \Description{Figure showing how NDCG varies with respect to search result list length $k$ and attack probability $pr$.}
        \caption{NDCG}
    \end{subfigure}
    \begin{subfigure}[t]{0.37\textwidth}
    \centering
        \includegraphics[width=\textwidth,keepaspectratio]{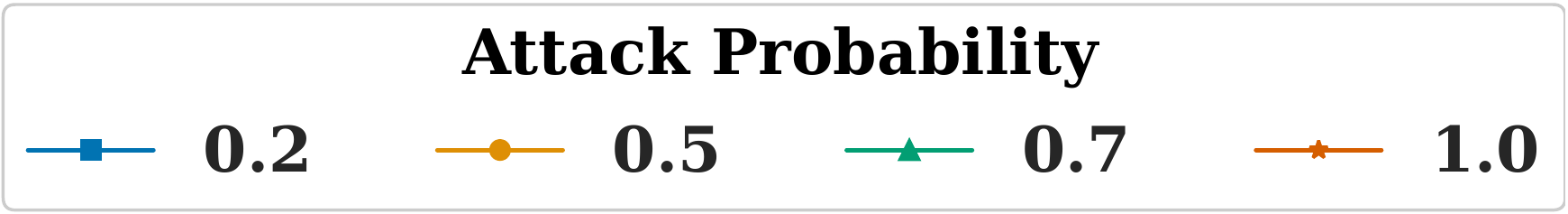}
        \Description{Legend for the above figures, with color codes lines for four increasing attack probabilities between 0.2 and 1.0.}
    \end{subfigure}
    \caption{Attack effectiveness as a function of attack probability $pr$ and list length $k$. Higher $\eta$ is a more effective attack, \ie the search results are more favorable to light-skinned men. Unfairness increases as $pr$ increases, yet there is almost no impact on ranking quality (NDCG). As $k$ increases skew is less impacted but attention is impacted somewhat more.}
    \label{fig:pandk}
     \vspace*{10pt}
\end{figure*}

\begin{figure*}[t]
    \begin{subfigure}[t]{0.3\textwidth}
        \centering
        \includegraphics[width=\columnwidth,keepaspectratio]{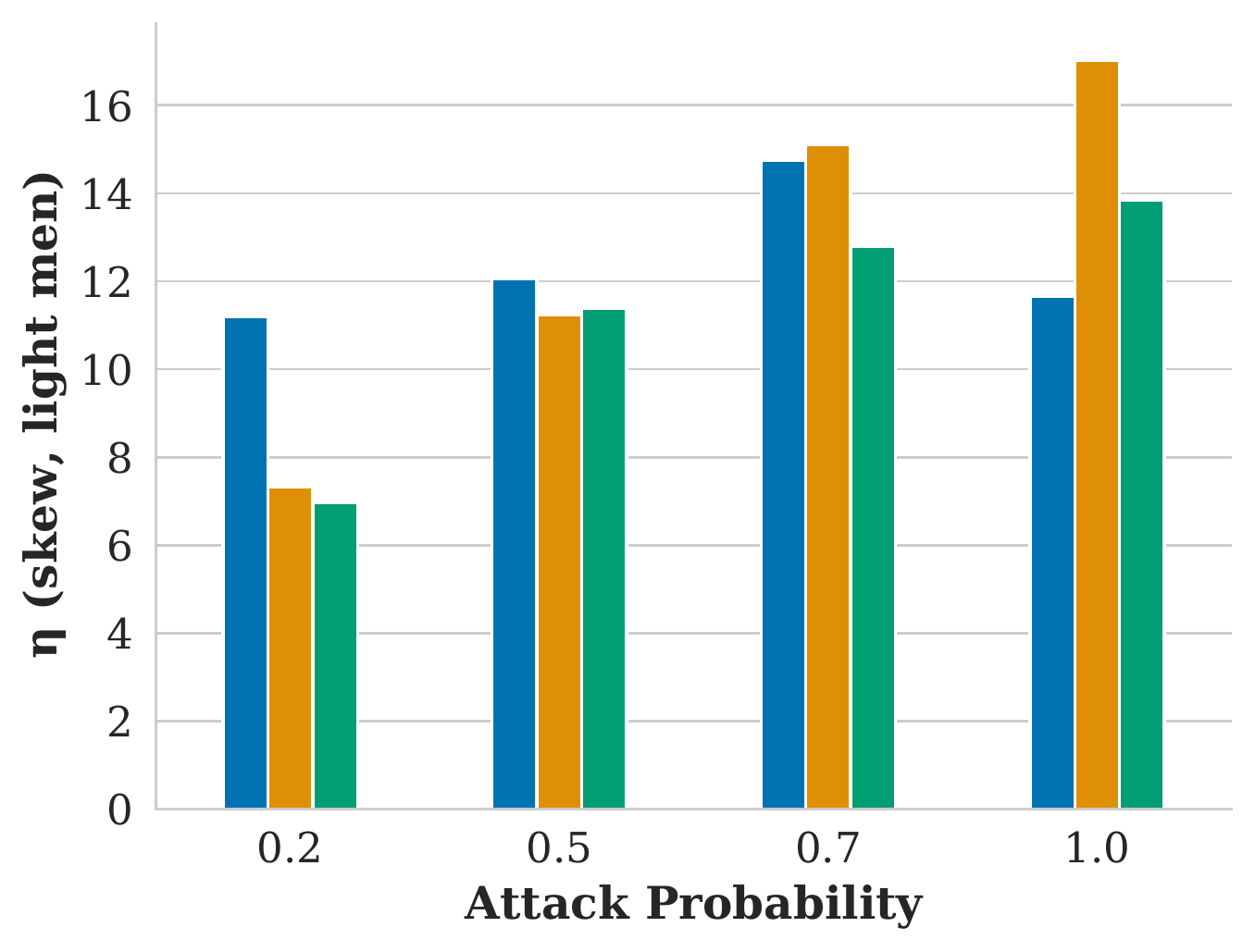}
        \Description{Figure showing how skew varies with respect to attack probability $pr$ and FMMR embedding model.}
        \caption{Skew}
    \end{subfigure}
    \begin{subfigure}[t]{0.3\textwidth}
        \centering
        \includegraphics[width=\columnwidth,keepaspectratio]{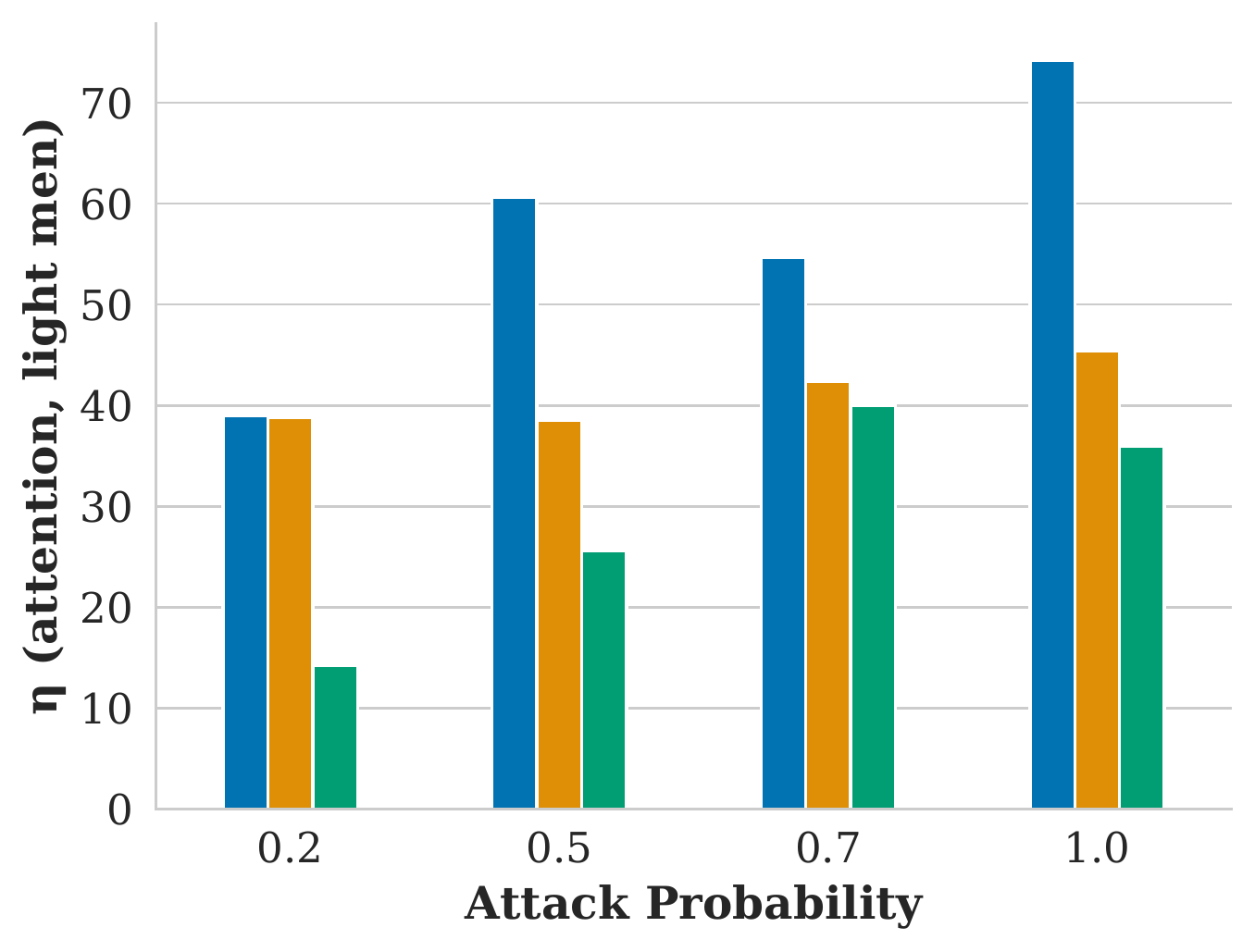}
        \Description{Figure showing how attention varies with respect to attack probability $pr$ and FMMR embedding model.}
        \caption{Attention}
    \end{subfigure}
    \begin{subfigure}[t]{0.3\textwidth}
        \centering
        \includegraphics[width=\columnwidth,keepaspectratio]{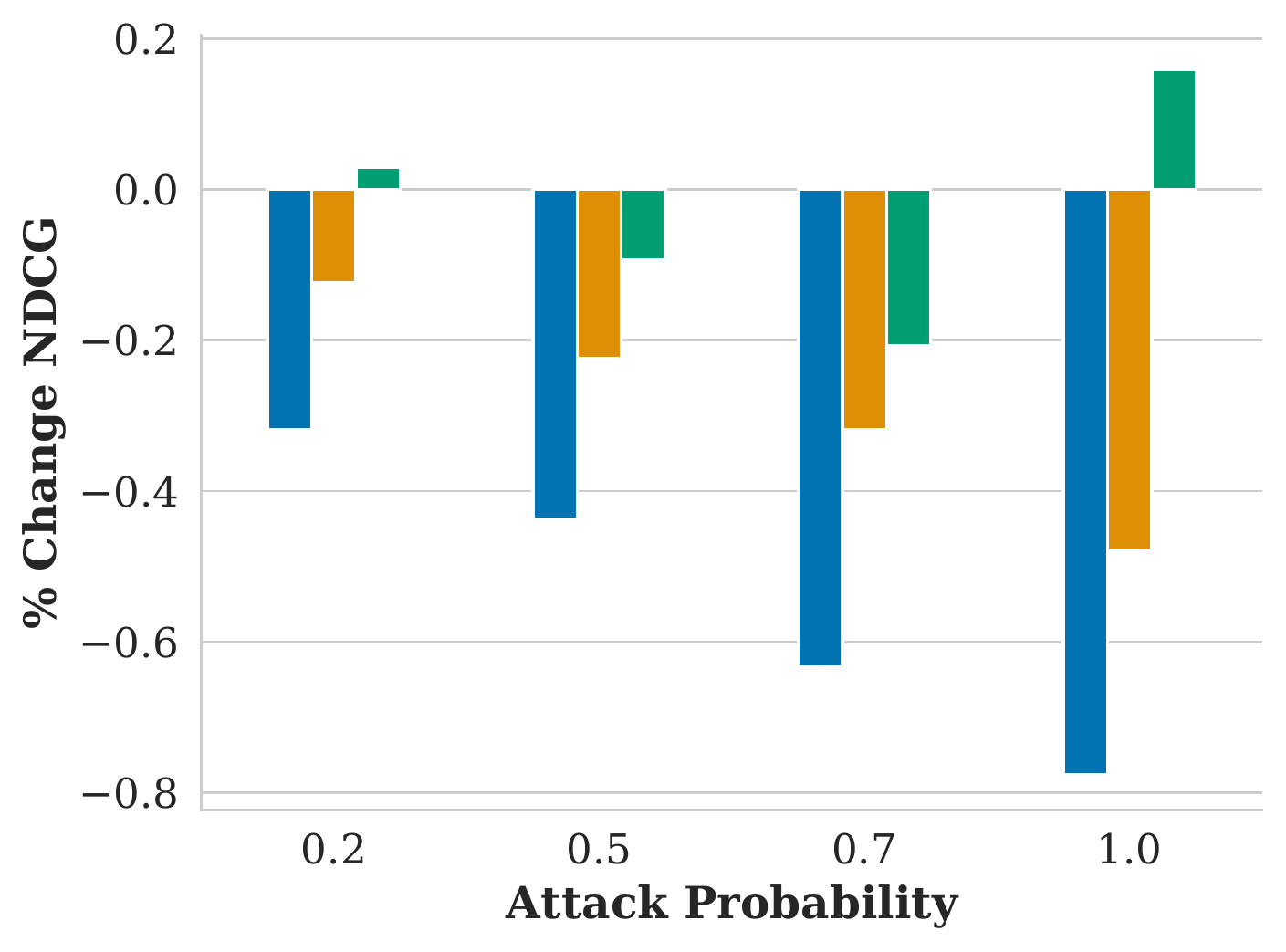}
        \Description{Figure showing how NDCG varies with respect to attack probability $pr$ and FMMR embedding model.}
        \caption{NDCG}
    \end{subfigure}
    \begin{subfigure}[t]{0.42\textwidth}
    \centering
        \includegraphics[width=\textwidth,keepaspectratio]{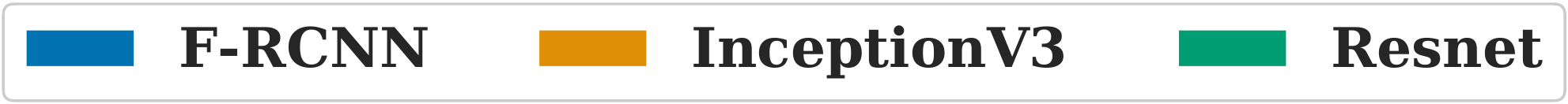}
        \Description{Legend for the above figures, with color coded bars representing attack effectiveness against three embedding algorithms: F-RCNN, InceptionV3, and Resnet.}
    \end{subfigure}
    \caption{Attack effectiveness is stable when the model used for the FMMR embedding is changed. ResNet embeddings are slightly more robust to attack and F-RCNN are slightly less robust. Interestingly, the ResNet's robustness is in spite of it having the most similar model architecture to FairFace.}
    \label{fig:embeddingfn}
    % \vspace*{15pt}
\end{figure*}

\subsection{Top \textit{k} and \textit{pr}}

We begin by evaluating the impact of our attacks as we vary the length of the top list $k$ and the fraction of images in the query list under adversarial control $pr$, plotted in \autoref{fig:pandk}. 

Varying $pr$ has the expected effect: as the adversary has more control over the image database, attacks become more effective, \ie $\eta$ for skew and attention increase. When the adversary is able to control 100\% of images in the query list, attacks are especially strong---increasing attention unfairness by over 50\% for some values of $k$. Even with only 20\% control, the adversary can increase attention unfairness by $\sim$30\%. Recall that $pr$ measures the fraction of each query list that is compromised, so as few as 35 images can be compromised at $pr=0.5$ (for the "Person eating Pizza" query). 

Varying $k$ also impacts ranking fairness. As $k$ increases, attention unfairness increases modestly and skew unfairness decreases. That skew unfairness decreases with $k$ indicates that the composition of items in the search results becomes fairer as the length of the list grows. However, our attack is able to cause FMMR to reorder the list such the top-most items remain unfair regardless of $k$, which is why attention unfairness exhibits less dependency on $k$.

Lastly, we observe that our attacks are stealthy. Regardless of $k$ or $pr$, NDCG never changes more than 0.7\%, meaning that our attack had effectively zero impact on search result relevance.

\begin{figure*}[t]
    \begin{subfigure}[t]{0.3\textwidth}
        \centering
        \includegraphics[width=\columnwidth,keepaspectratio]{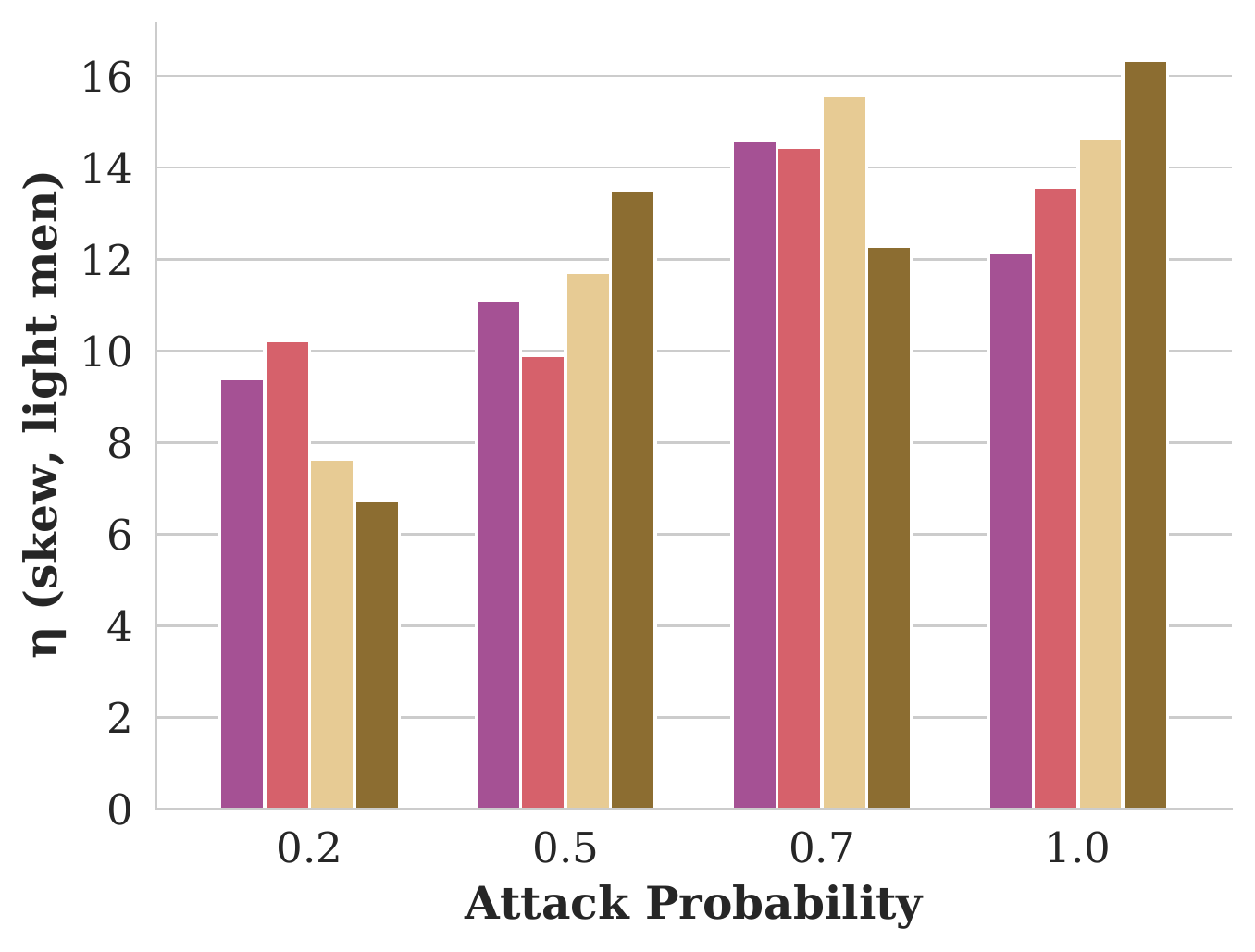}
        \Description{Figure showing how skew varies with respect to attack probability $pr$ and GAP training objective.}
        \caption{Skew}
    \end{subfigure}
    \begin{subfigure}[t]{0.3\textwidth}
        \centering
        \includegraphics[width=\columnwidth,keepaspectratio]{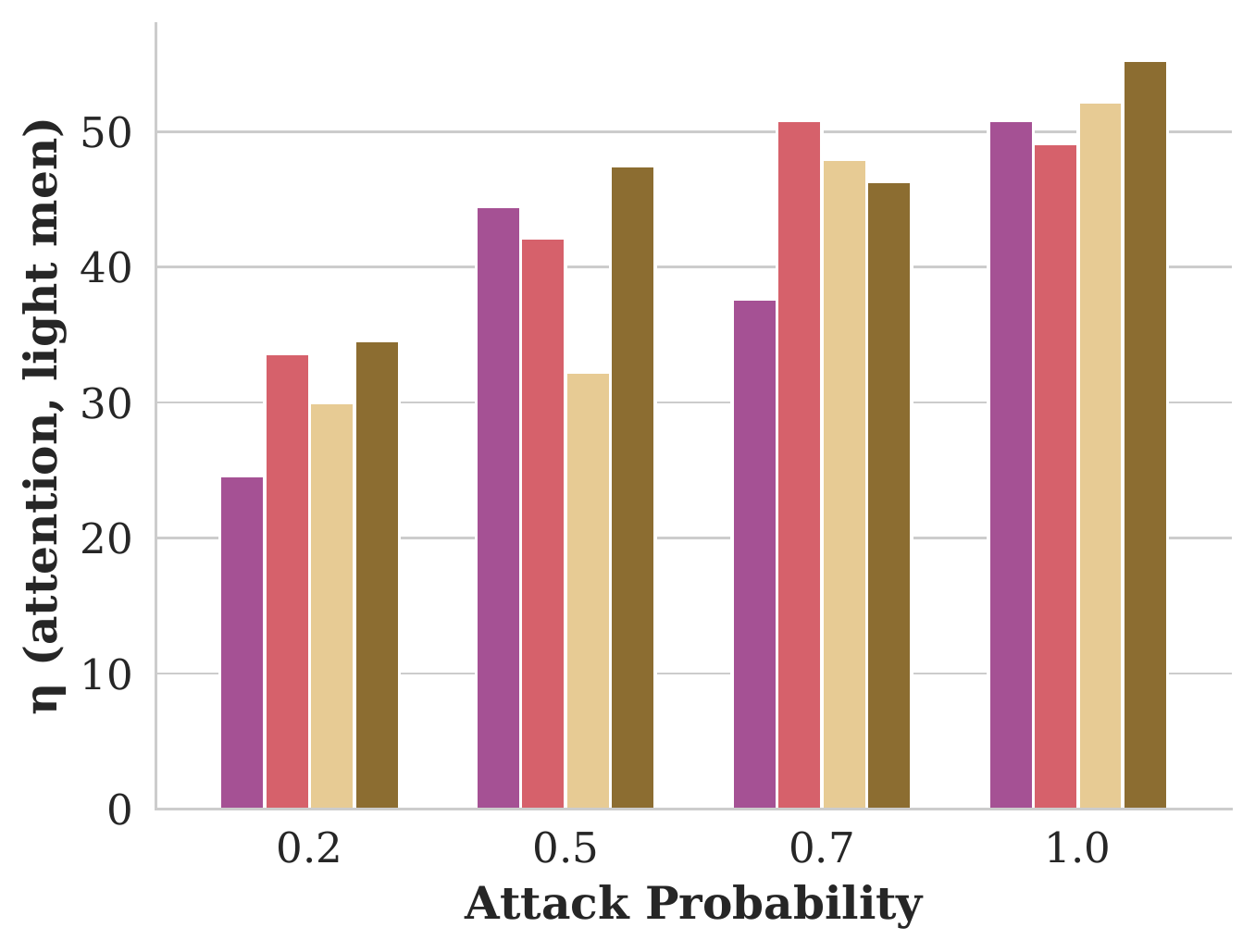}
        \Description{Figure showing how attention varies with respect to attack probability $pr$ and GAP training objective.}
        \caption{Attention}
    \end{subfigure}
    \begin{subfigure}[t]{0.3\textwidth}
        \centering
        \includegraphics[width=\columnwidth,keepaspectratio]{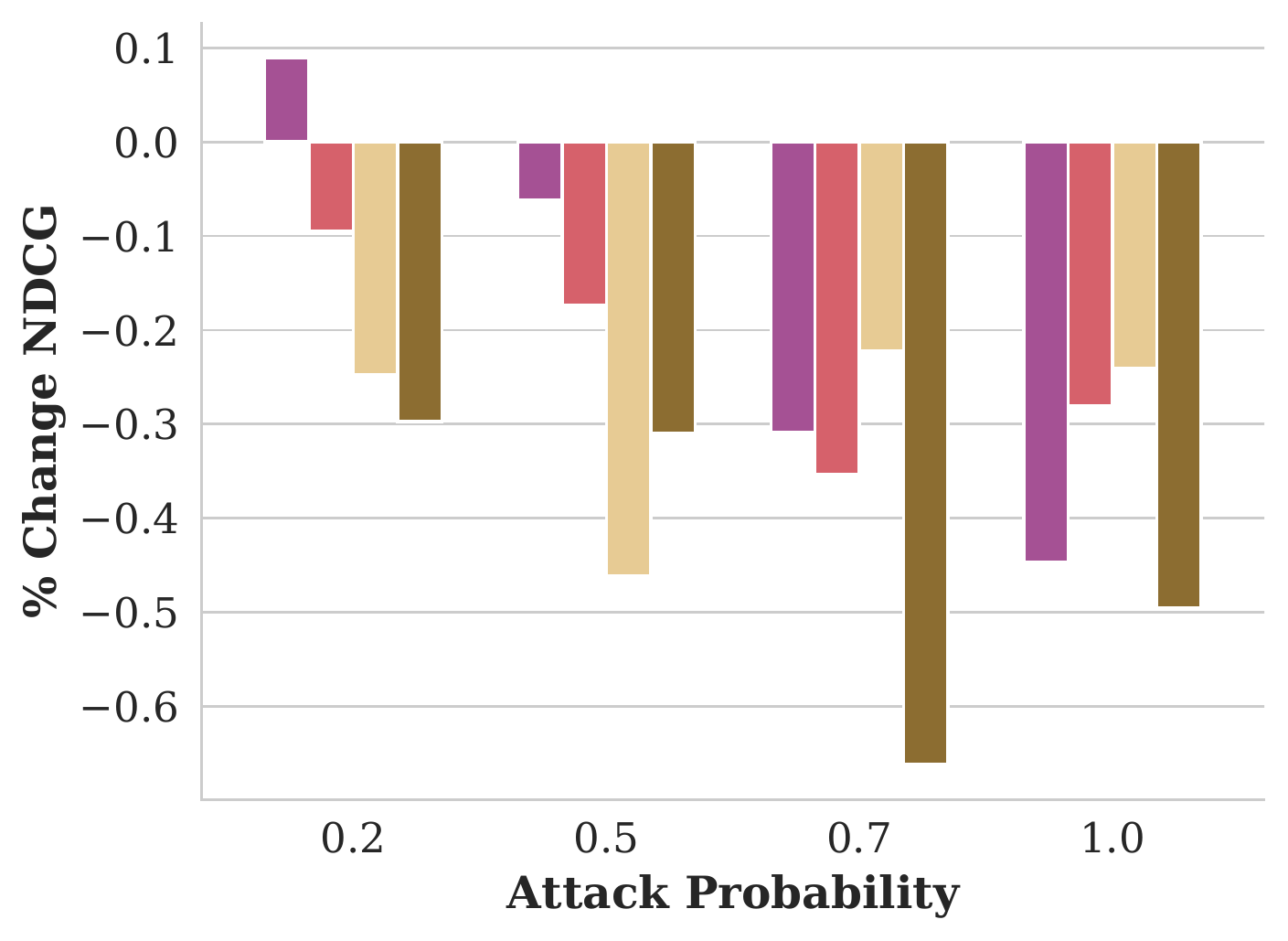}
        \Description{Figure showing how NDCG varies with respect to attack probability $pr$ and GAP training objective.}
        \caption{NDCG}
    \end{subfigure}
    \begin{subfigure}[t]{0.88\textwidth}
    \centering
        \includegraphics[width=\textwidth,keepaspectratio]{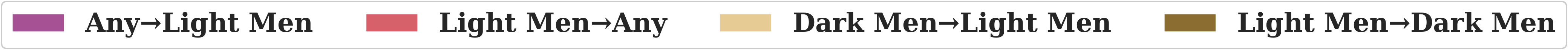}
        \Description{Legend for the above figures, with color coded bars corresponding to four GAP training objectives: perturbing all images to appear as light-skinned men, perturbing light-skinned men to appear as any other demographic group, perturbing dark-skinned men to appear as light-skinned men, and perturbing light-skinned men to appear as dark-skinned men.}
    \end{subfigure}
    \caption{Attack effectiveness is relatively stable when the GAP training objective is changed.}
    \label{fig:gapobjective}
    % \vspace{1em}
    \end{figure*}

    \begin{figure*}[t]
    \centering
    \begin{subfigure}[t]{0.75\textwidth}
        \centering
        \includegraphics[width=\columnwidth,keepaspectratio]{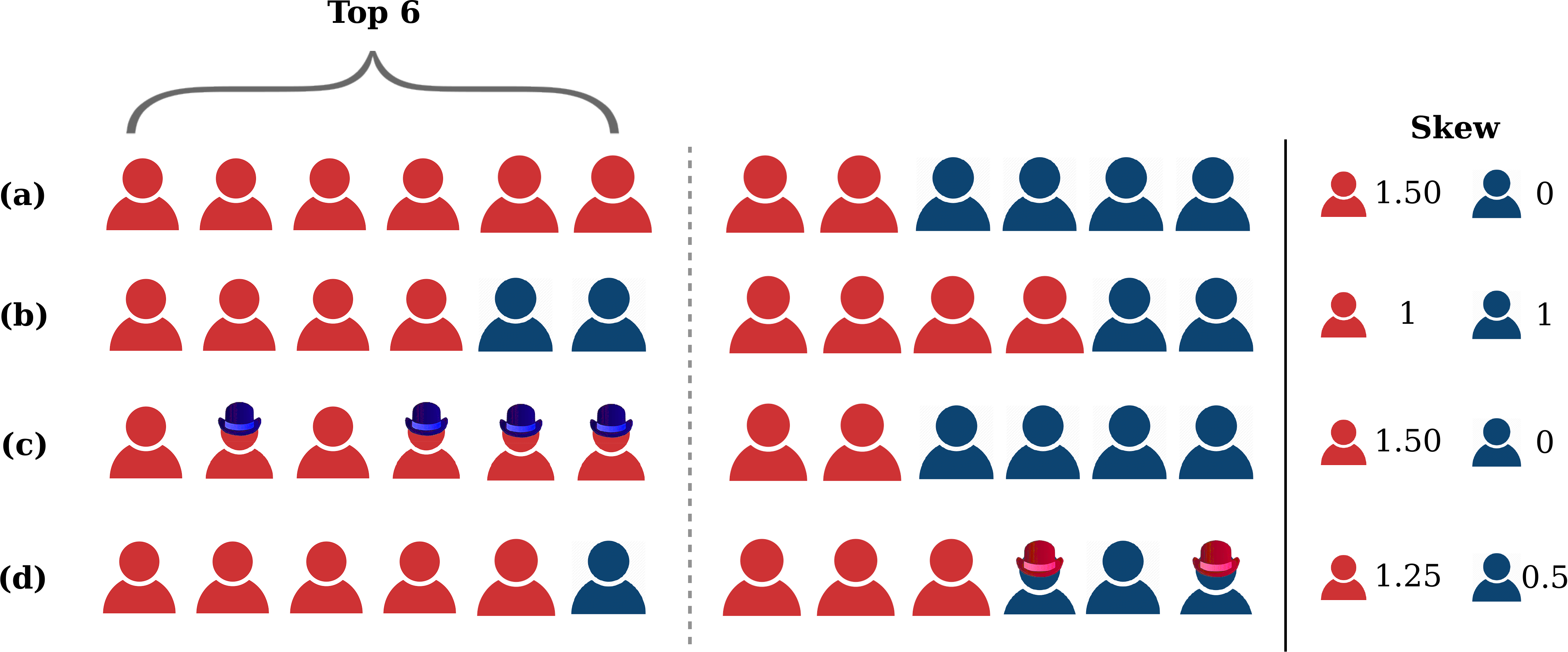}
        \Description{A graphical example of four sets of hypothetical search results. Each set of search results contains twelve individuals: eight light-skinned and four dark-skinned. The first set of results is the unfair baseline, with all dark-skinned people appearing at low ranks. The second set is group fair, with equal proportions of light and dark-skinned individuals in the top six results. The remaining two sets include perturbed images which cause the group fairness constraint to be violated.}
    \end{subfigure}
    \begin{subfigure}[t]{0.52\textwidth}
    \centering
        \includegraphics[width=\textwidth,keepaspectratio]{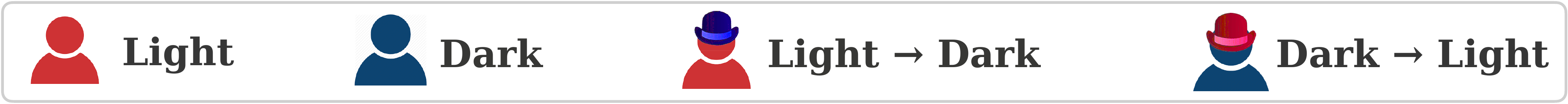}
        \Description{Legend for the above figure, describing the graphical figures for light and dark-skinned individuals, as well as perturbed and unperturbed individuals.}
    \end{subfigure}
    \caption{An example showing how incorrect group allocation in any direction always harms the minority group members in fair ranking. \textbf{(a)} shows a \textit{baseline} unfair list, with all people sorted by relevance to the query and no dark people in the top 6. \textbf{(b)} shows the fair ranking produced by FMMR, with the same proportion of light and dark people in the top 6 as the overall population. In \textbf{(c)}, light people's images are perturbed using a GAP so that half of them are grouped with dark people. FMMR moves the most relevant dark people into the top 6 to make the list fair, but in this case the most relevant ``dark'' people are really light skinned. In \textbf{(d)}, half of the dark people are perturbed using a GAP to be grouped as light people. To FMMR, this appears to reduce the overall population of dark people, so it only needs to move one dark person into the top 6 to make the list proportionally fair. Note that if all light people were grouped as dark or all dark people were grouped as light, the ranking would remain the unfair baseline shown in \textbf{(a)}.}
    \label{fig:minority}
    \vspace*{8pt}
\end{figure*}

\subsection{Choice of Training Objective}

We evaluate our attack's impact on fairness with four CGAP models: one that misclassifies Dark Men as Light Men, one for misclassifying Light Men as Dark Men, and relaxed CGAP models that misclassify all people as Light Men and all Light Men as other groups. We show these attacks' effectiveness in \autoref{fig:gapobjective}.

Each of these attacks performs similarly well at harming fairness in terms of skew and attention, and remaining stealthy in terms of NDCG. One surprising observation is that misclassifying Dark Men as Light Men performs similarly to the exact opposite attack: in both cases, Light Men end up with an significant, unfair advantage. We explain this seeming contradiction with an example in \autoref{fig:minority}. In essence, using a GAP to misclassify people from a minority group into the majority group reduces the minority group's overall share of the population. Since group fairness in this case is based on the overall population distribution, this causes FMMR to rerank fewer minority group members into the top of the search results.

Based on the results in \autoref{fig:gapobjective}, it appears that there is no way to advantage a minority group with our attacks.

\subsection{Choice of Attack Training Algorithm}

We measure our attacks' effectiveness when the GAP models are trained on Deepface and FairFace demographic inference models. We observe that attack effectiveness is largely independent of the choice of inference model, and all attacks remain stealthy. We defer a plot of the results to the supplementary material, in \autoref{fig:demoinf}.

% CBW --- This is an interesting question to explore, but our answer is totally unsatisfying, so cutting for now.
%It is unclear why the robustness of the embedding models varies. For example, it is possible that decisions made by F-RCNN are more similar to the demographic inference models upon which our attack was trained, leading to its low robustness.

%FairFace seems to result in attacks that are slightly more harmful to attention, and have slightly higher impact on NDCG (although still small). Meanwhile, Deepface results in slightly higher skew.

\begin{figure*}[t]
    \begin{subfigure}[t]{0.3\textwidth}
        \centering
        \includegraphics[width=\columnwidth,keepaspectratio]{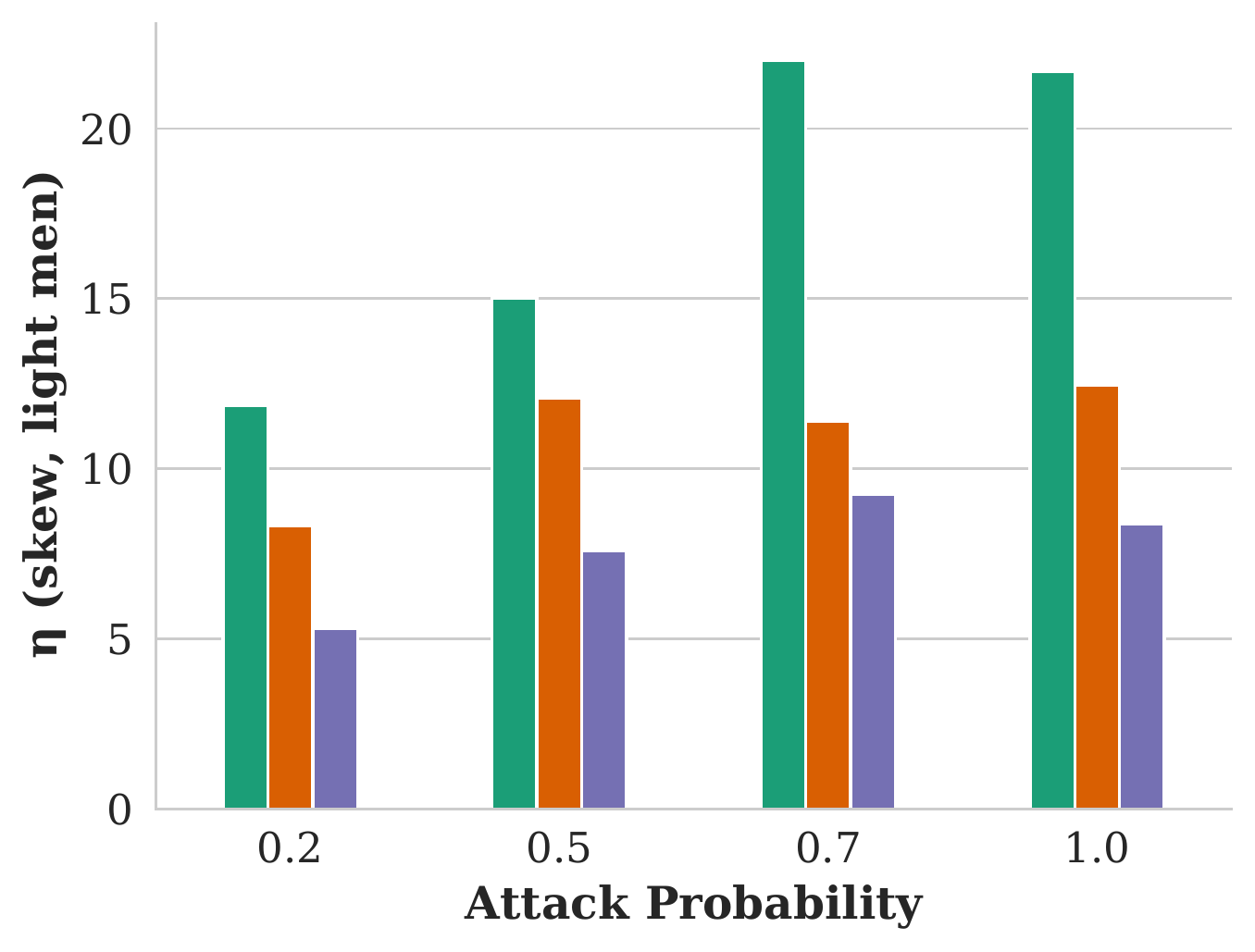}
        \Description{Figure showing how skew varies with respect to attack probability $pr$ and search query.}
        \caption{Skew}
    \end{subfigure}
    \begin{subfigure}[t]{0.3\textwidth}
        \centering
        \includegraphics[width=\columnwidth,keepaspectratio]{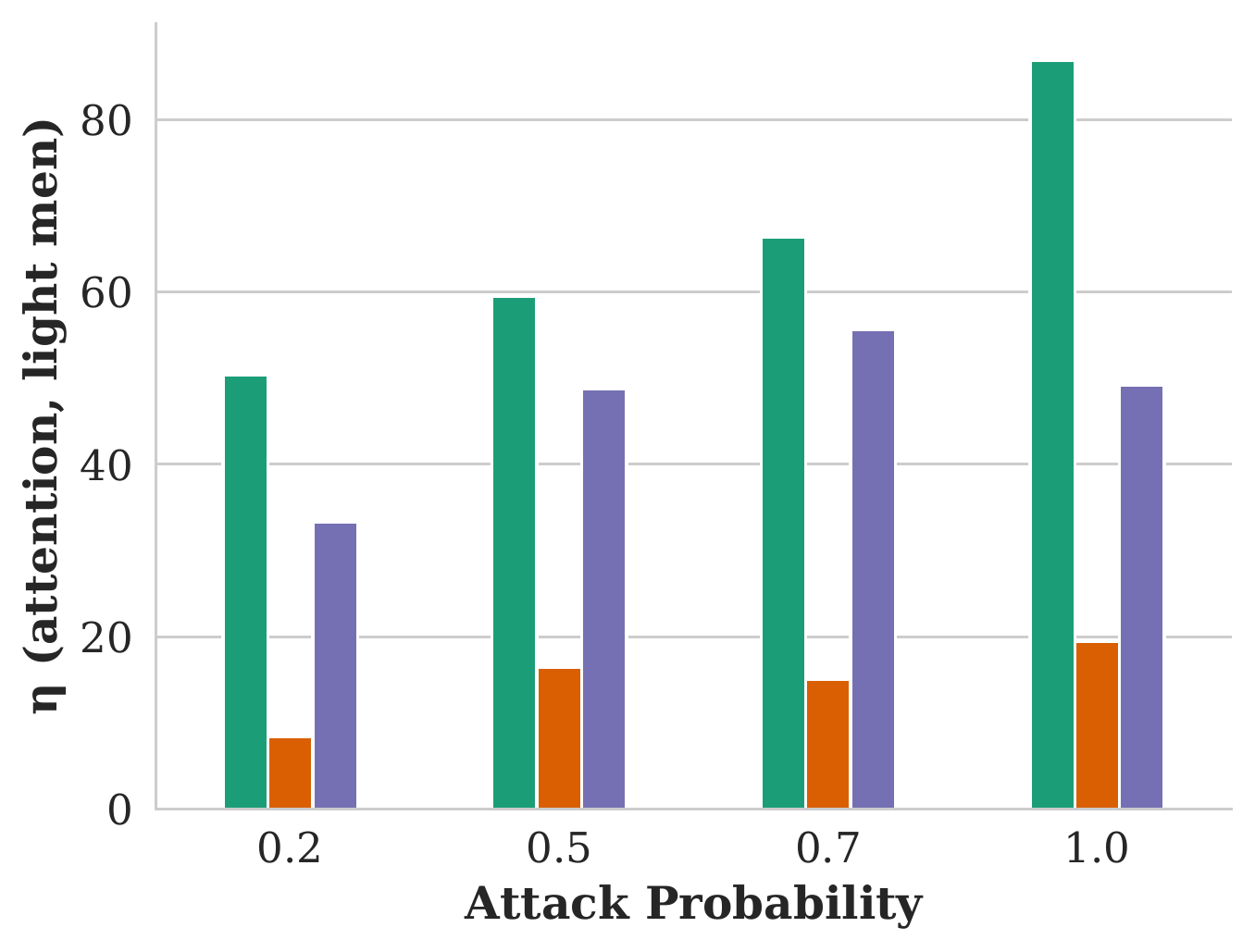}
        \Description{Figure showing how attention varies with respect to attack probability $pr$ and search query.}
        \caption{Attention}
    \end{subfigure}
    \begin{subfigure}[t]{0.3\textwidth}
        \centering
        \includegraphics[width=\columnwidth,keepaspectratio]{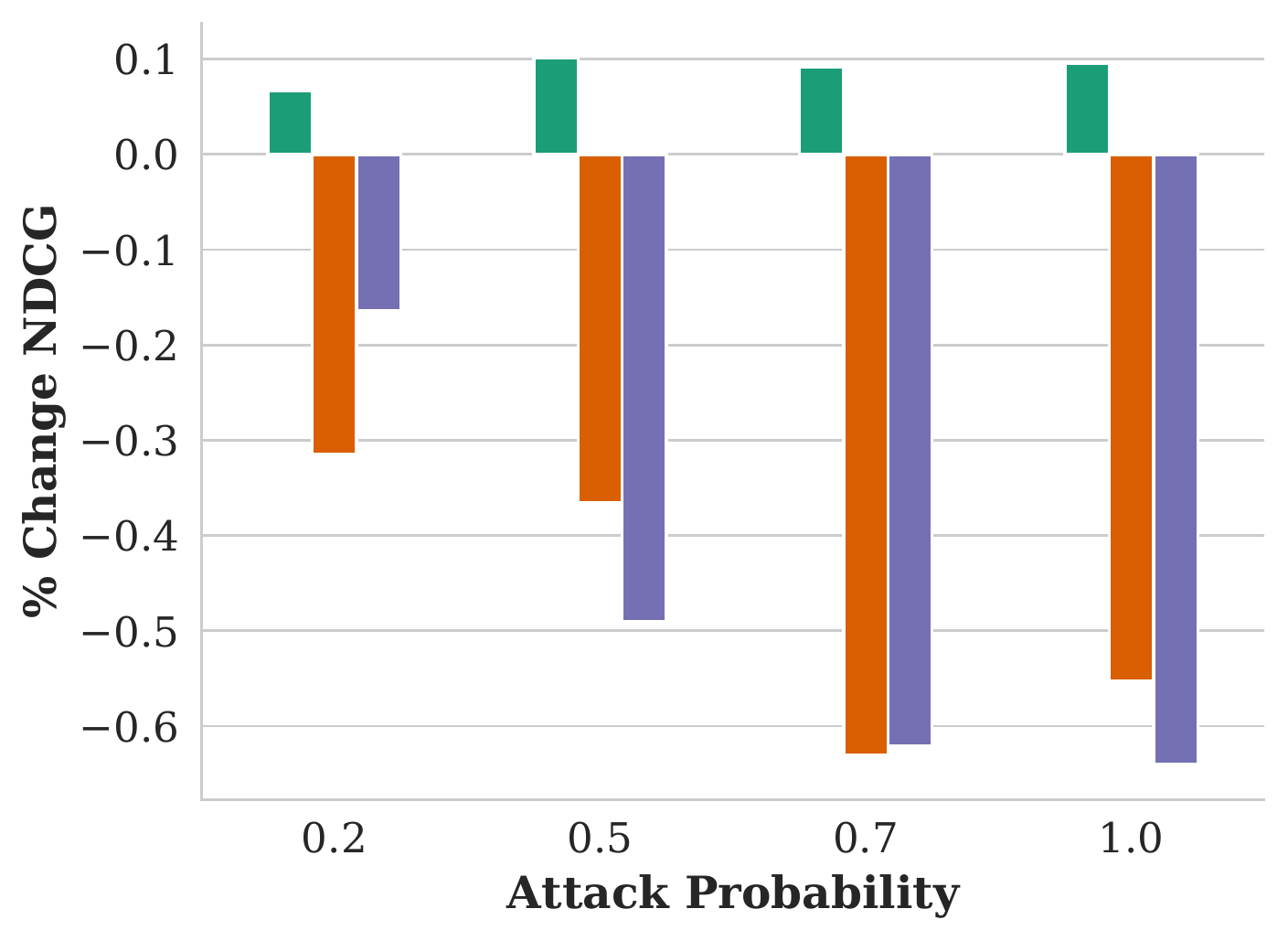}
        \Description{Figure showing how NDCG varies with respect to attack probability $pr$ and search query.}
        \caption{NDCG}
    \end{subfigure}
    \begin{subfigure}[t]{0.88\textwidth}
    \centering
        \includegraphics[width=\textwidth,keepaspectratio]{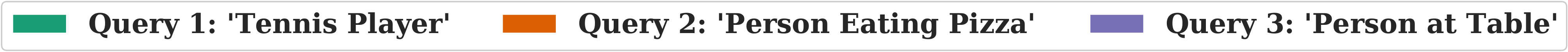}
        \Description{Legend for the above figures, with color coded bars for three search queries: ``tennis player'', ``person eating pizza'', and ``person at table''.}
    \end{subfigure}
    \caption{Attacks are effective against all three of our queries, but the effectiveness varies in relation to the underlying population and utility score distributions (see \autoref{fig:circleplot}).}
    \label{fig:query}
    %\vspace*{-8pt}
\end{figure*}

\subsection{Choice of Query}

Lastly, we examine the effectiveness of our attacks against three different queries and plot the results in \autoref{fig:query}. We observe that all attacks were successful, but that effectiveness varies by query. The differences in attack effectiveness are explained by the underlying distributions of population and utility scores (see \autoref{fig:circleplot}). The ``tennis'' results exhibit the most unfairness post-attack because they were most unfair to begin with, \ie the difference in utility scores between Light and Dark skinned people was greatest in the ``tennis'' results as compared to the other queries. In contrast, the ``pizza'' results exhibit the most robustness to attack in terms of attention because these were the only results among the queries where minority people had higher utility scores than majority people in the baseline results (\autoref{fig:circleplot}).

%---------------------------------------------------------------------------------------------
%---------------------------------------------------------------------------------------------
%---------------------------------------------------------------------------------------------
%---------------------------------------------------------------------------------------------

\section{Discussion}

In this study, we develop a novel, adversarial ML attack against fair ranking algorithms, and use fairness-aware text-to-image retrieval as a case study to demonstrate our attack's effectiveness. Unfortunately, we find that our attack is very successful at subverting the fairness algorithm of the search engine---across an extensive set of attack variations---while having almost zero impact on search result relevance.

Although we present a single case study, we argue that our attack is likely to generalize. We adopt a strong threat model and demonstrate that our attacks succeed even when the attacker cannot poison training data, access the victim's whole image corpus, or know what models are used by the victim. Thus, our attack is highly likely to succeed in cases where the threat model is more relaxed, \eg when the fairness algorithm used by the victim is known.

%Similarly, although we only demonstrate attacks against neural image models, we hypothesize that our attack could be adapted to work against IR systems that leverage other highly-parameterized models.

Alarmingly, our work shows that an adversary can attack a fairness algorithm like FMMR \textit{even when it does not explicitly rely on demographic inference}. Thus, it is highly likely that our attack will also succeed against any ranking algorithm that does rely on a demographic inference model, even if that model is highly accurate. We explore this possibility to the best of our ability in \autoref{sec:detconstsort}.

%Prior work~\cite{ghosh2021fair} has shown that fair ranking algorithms that rely on demographic labels can perform poorly when these labels are generated with imperfect demographic inference algorithms.  This demonstrates that avoiding the use of demographic inference cannot be a simple ``bug fix'' for the issues raised in our study or by \citet{ghosh2021fair}.

We hope that this research will raise awareness and spur further research into vulnerabilities in fair algorithms. Our results highlight how, in the absence of safeguards, fairness interventions can potentially be weaponized by malicious parties as a tool of oppression. In the absence of fair ML development methods and algorithms that are robust to adversarial attacks, it may not be possible for policymakers to safely mandate the use of fair ML algorithms in practice.

We believe that future work is needed to develop more robust fair ML interventions. We adopt a broad view of possible mitigations, spanning from value sensitive design~\cite{vsd} methods that help developers preemptively identify attack surface and plan defenses~\cite{security-cards}, to models that are hardened against adversarial perturbation techniques~\cite{akhtar2018defense, goodfellow2014explaining}, to auditing checklists~\cite{raji-2020-fat} and tools that help developers notice and triage attacks.

Above all, this work highlights that achieving demographic fairness requires high-quality demographic data~\cite{andrus2021measure}. Allowing an adversary to influence demographic meta-data is the underlying flaw that enables our attack to succeed. Demographic data may be sourced from data subjects themselves, with full knowledge and consent, or from human labelers~\cite{basu-2020-lighthouse}, with the caveat that these labels themselves will need to be de-biased~\cite{zhao2021captionbias}.

\subsection{Limitations}
\label{sec:limitation} 

Our study has a number of limitations. First, our analysis is limited to two discrete racial and two discrete gender categories. Although our CGAP attack could be tailored to select any group, it is unclear how well our attack would perform in situations with $>4$ discrete protected groups, groups with continuous attributes, people with multiple or partial group memberships, or with population distributions that varied significantly from our dataset. Second, while our dataset is sufficiently large to demonstrate our attack, it is smaller than the databases that real-world image search engines retrieve from. Third, our proof-of-concept was tuned to attack FairFace and Deepface. It is unclear how well our CGAP attack would generalize to other models or real-world deployed systems. Fourth, as we observe in \autoref{fig:gapobjective}, our attack is only successful at generating unfairness in favor of already-advantaged groups. While this is a limitation, it in no way diminishes the potential real-world harm our attack could inflict on marginalized populations. Finally, as shown in \autoref{fig:query}, our attack's effectiveness varies by query. In real world scenarios, an attacker could mitigate this to some extent by devoting more of their resources towards perturbing images that are relevant to high-value queries. It is unclear how much the attackers' effort would need to vary in practice given that we make no attempt to attack deployed search engines.

\subsection{Ethics}

In this work we present a concrete attack against a fair ranking system. Like all adversarial attack research, our methods can potentially be misused by bad actors. However, this also necessitates our research, since documenting vulnerabilities is the first step towards mitigating them. To the best of our knowledge, with the exception of Shopify and LinkedIn, few services are known to employ fair ranking systems in practice, meaning there currently exists a window of opportunity to preemptively identify attacks, raise awareness, and deploy mitigations.

Prior work on adversarial ML attacks against fairness made their source code publicly available~\cite{nanda2021fairness}. However, because attack tools are dual-use, we have opted to take a more conservative approach: we will only share source code with researchers from (1) research universities (\eg as identified by taxonomies like the Carnegie Classification) and (2) companies that develop potentially vulnerable products. Given that our attack can be used for legitimate, black-box algorithm auditing purposes, we opt to restrict who may access our source code rather than the uses it may be put towards. In our opinion, this process will facilitate follow-up research, mitigation development, and algorithm auditing without supplying bad actors with ready-made attack tools.

Like many works in the computer vision field, we rely on images with crowdsourced and inferred demographic labels. Both processes have been criticized for their lack of consent~\cite{gies-2020-sieds}, the way they operationalize identity~\cite{scheuerman-2020-cscw}, and the harm they may cause through mis-identification~\cite{bennett-2021-chi}. These problems reinforce the need for high-quality, consensual demographic data as a means to improve ethical norms and defend against adversarial ML attacks.

% Reviewer 3 Comment
% Who are the “well-known institutions and companies”? And why is being “well-known” the only criteria, versus other criteria? The authors could require some kind of agreement that the recipient must sign, committing to not use the code to attack a system (perhaps except for testing purposes or with the owner/operator’s consent?)

\begin{acks}
\paragraph{Funding/Support.} This work was supported by a 2019 Sloan Fellowship and \grantsponsor{1}{NSF}{https://www.nsf.gov/} grants \grantnum{1}{IIS-1553088} and \grantnum{1}{IIS-1910064}. Any opinions, findings, conclusions, or recommendations expressed herein are those of the authors and do not necessarily reflect the views of the funders.

The legal department of Google participated in the review and approval of the manuscript. They asked that the authors add additional examples of image search engines besides Google Image Search, and we complied. Aside from the authors, Google had no role in the design and conduct of the study; access and collection of data; analysis and interpretation of data; or preparation of the manuscript. 

The authors thank Jane L. Adams, Michael Davinroy, Jeffrey Gleason, and the anonymous reviewers for their comments.
\end{acks}

% \balance
\bibliographystyle{ACM-Reference-Format}
\bibliography{ref}

\newpage

\appendix

\section{Supplementary Material}
\label{sec:supplementary}

\begin{figure*}[t]
    \begin{subfigure}[t]{0.3\textwidth}
        \centering
        \includegraphics[width=\columnwidth,keepaspectratio]{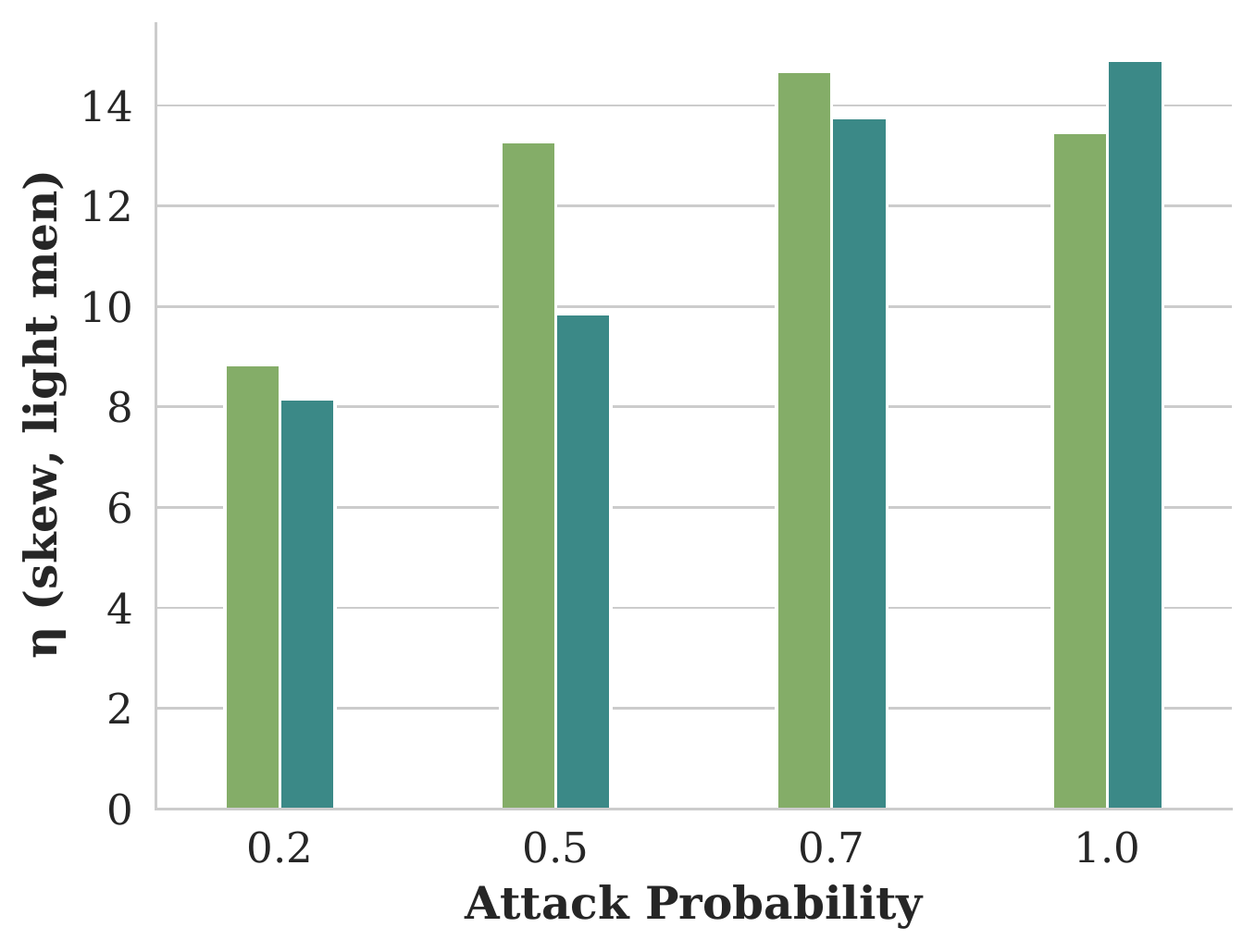}
        \Description{Figure showing how skew varies with respect to attack probability $pr$ and demographic inference model used to train the GAP.}
        \caption{Skew}
    \end{subfigure}
    \begin{subfigure}[t]{0.3\textwidth}
        \centering
        \includegraphics[width=\columnwidth,keepaspectratio]{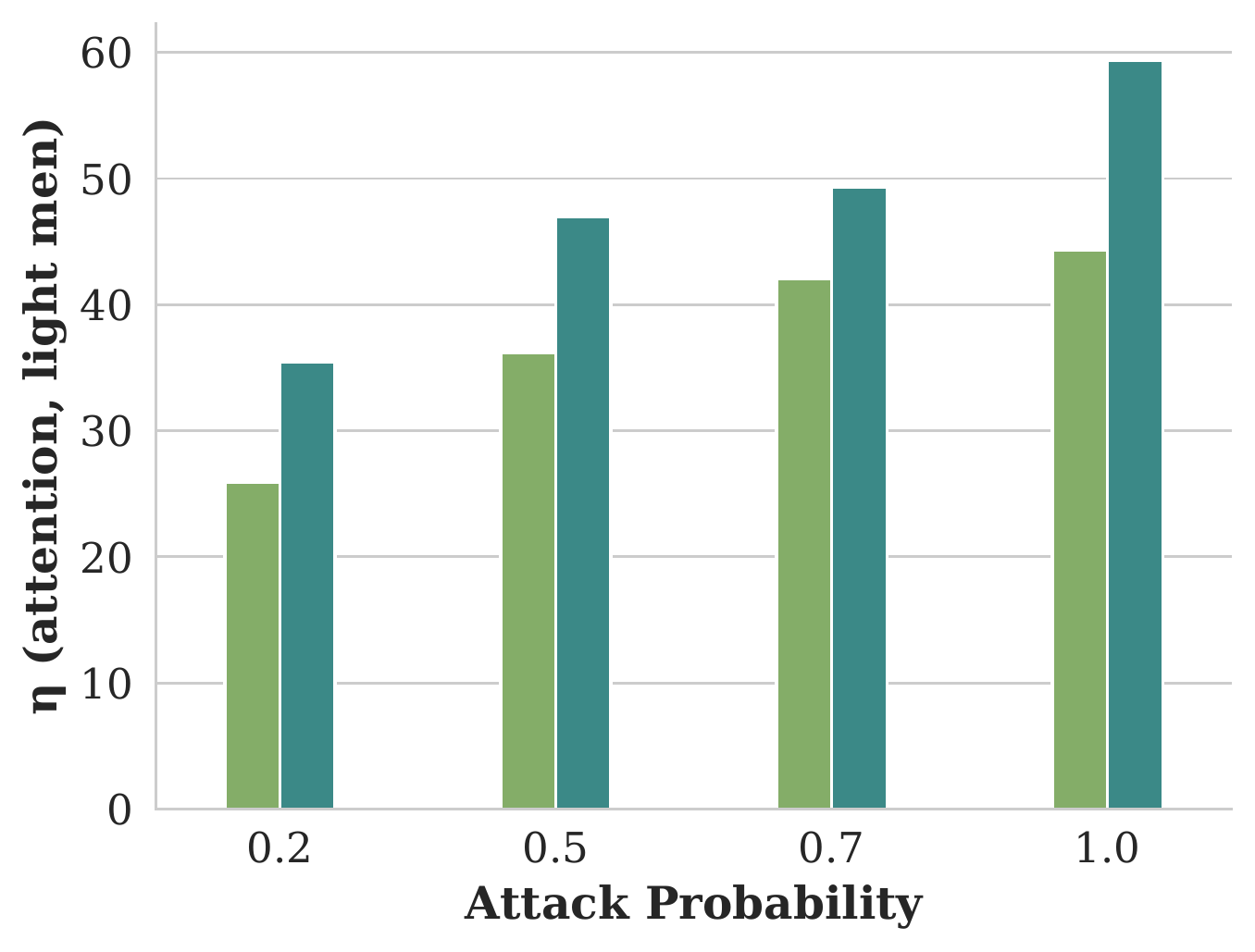}
        \Description{Figure showing how attention varies with respect to attack probability $pr$ and demographic inference model used to train the GAP.}
        \caption{Attention}
    \end{subfigure}
    \begin{subfigure}[t]{0.3\textwidth}
        \centering
        \includegraphics[width=\columnwidth,keepaspectratio]{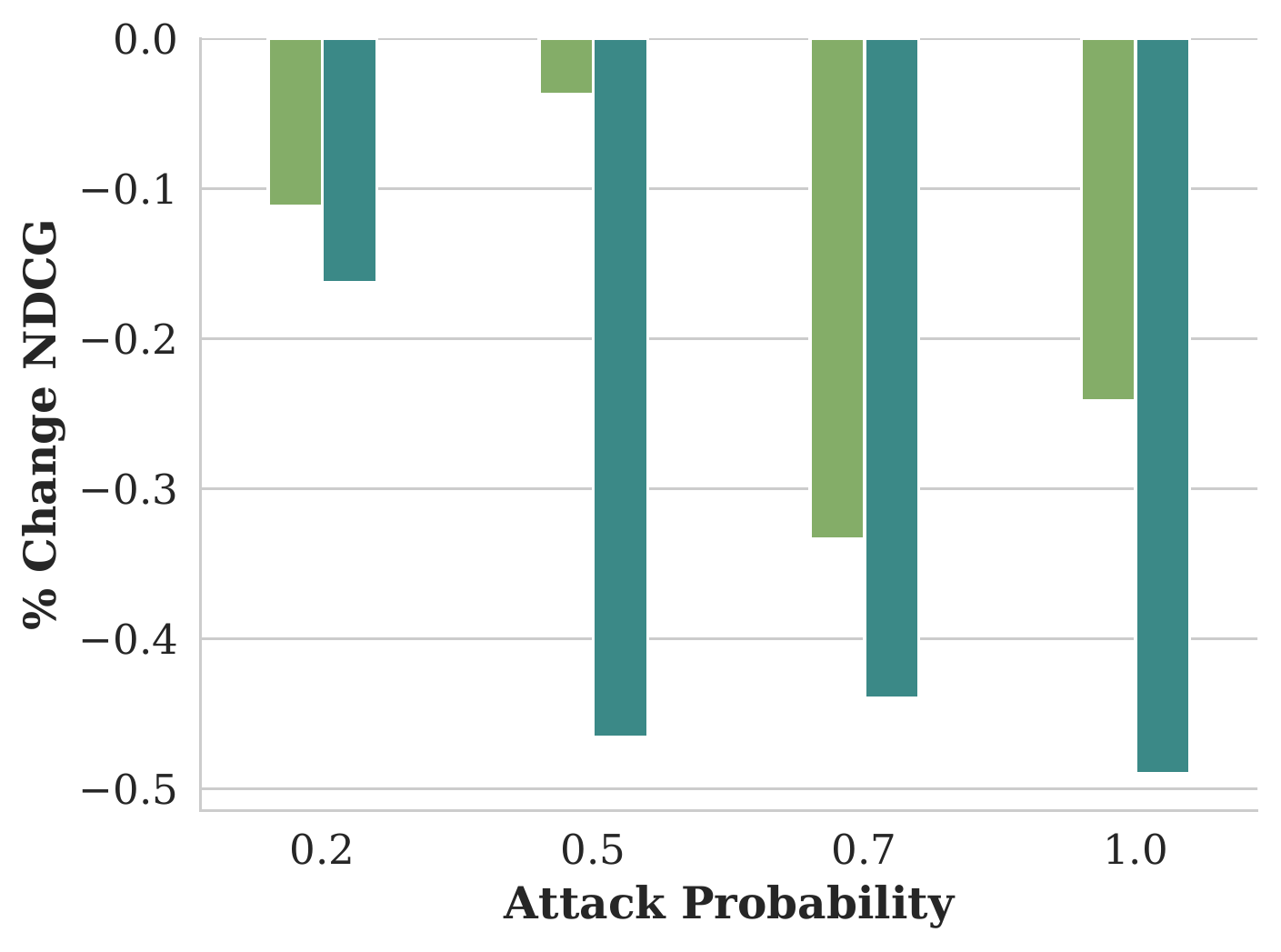}
        \Description{Figure showing how NDCG varies with respect to attack probability $pr$ and demographic inference model used to train the GAP.}
        \caption{NDCG}
    \end{subfigure}
    \begin{subfigure}[t]{0.28\textwidth}
    \centering
        \includegraphics[width=\textwidth,keepaspectratio]{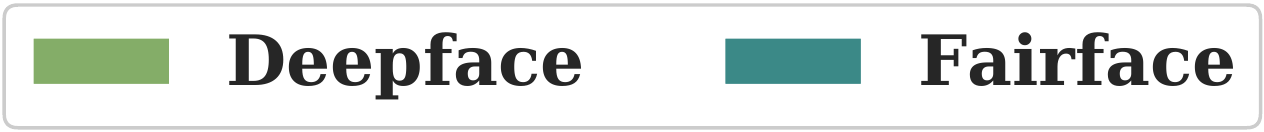}
        \Description{Legend for the above figures, with color coded bars for two demographic inference models: Deepface and Fairface.}
    \end{subfigure}
    \caption{GAP models trained on different demographic inference algorithms offer similar attack effectiveness.}
    \label{fig:demoinf}
\end{figure*}

\begin{figure*}[t]
    \begin{subfigure}[t]{0.3\textwidth}
        \centering
        \includegraphics[width=\columnwidth,keepaspectratio]{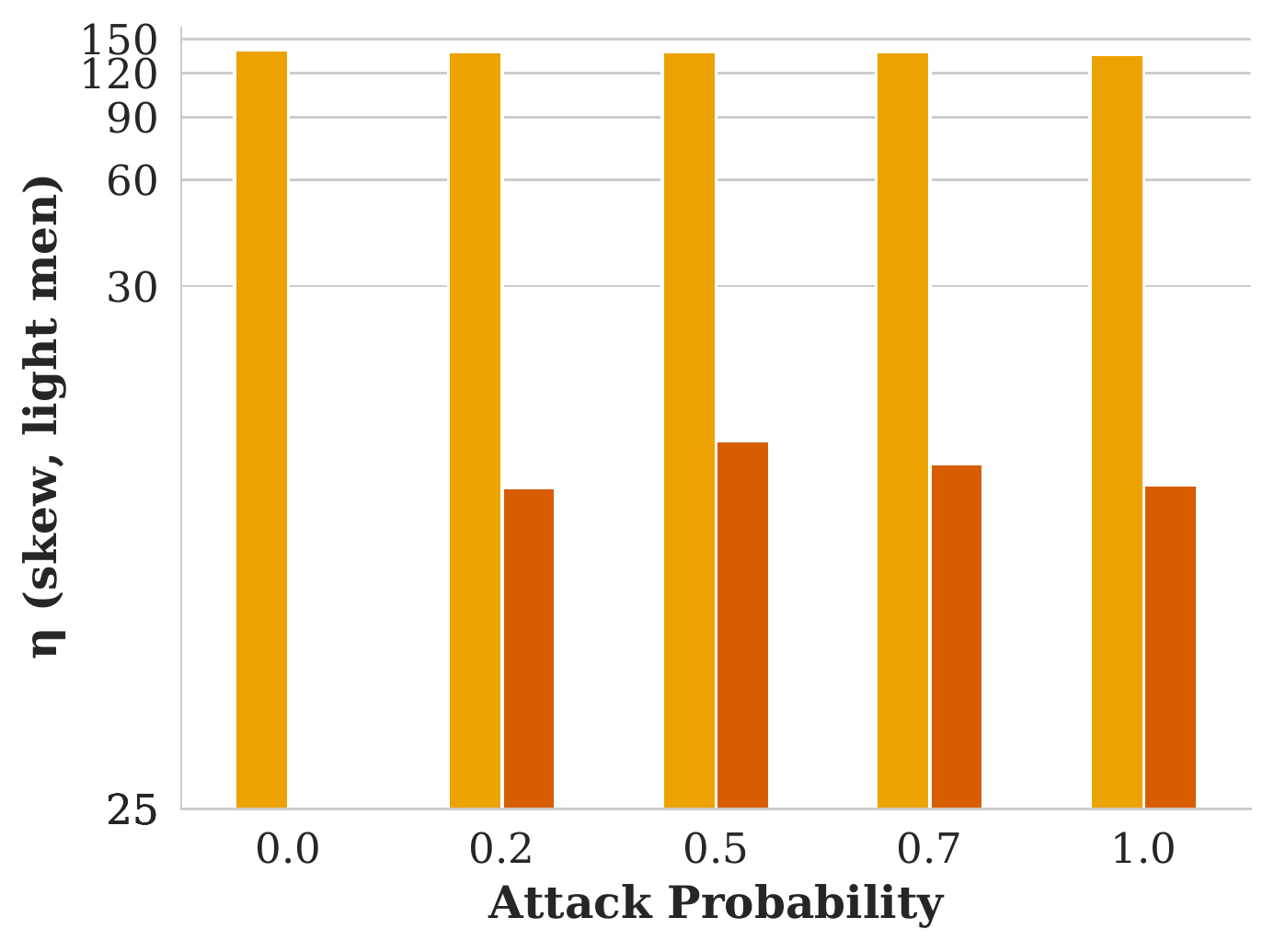}
        \Description{Figure showing how skew varies with respect to attack probability $pr$ and fair re-ranking algorithms used by the image search engine.}
        \caption{Skew}
    \end{subfigure}
    \begin{subfigure}[t]{0.3\textwidth}
        \centering
        \includegraphics[width=\columnwidth,keepaspectratio]{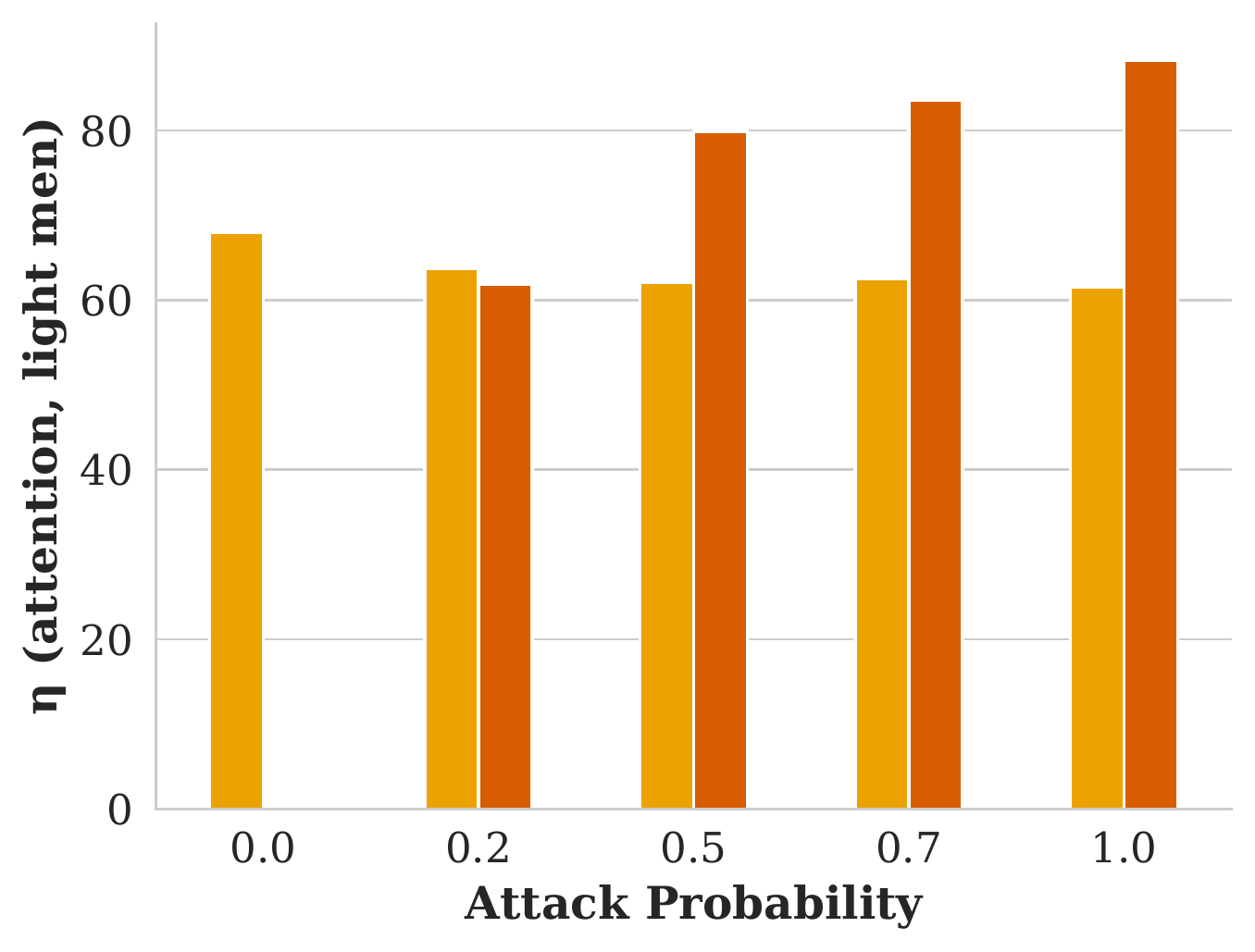}
        \Description{Figure showing how attention varies with respect to attack probability $pr$ and fair re-ranking algorithms used by the image search engine.}
        \caption{Attention}
    \end{subfigure}
    \begin{subfigure}[t]{0.3\textwidth}
        \centering
        \includegraphics[width=\columnwidth,keepaspectratio]{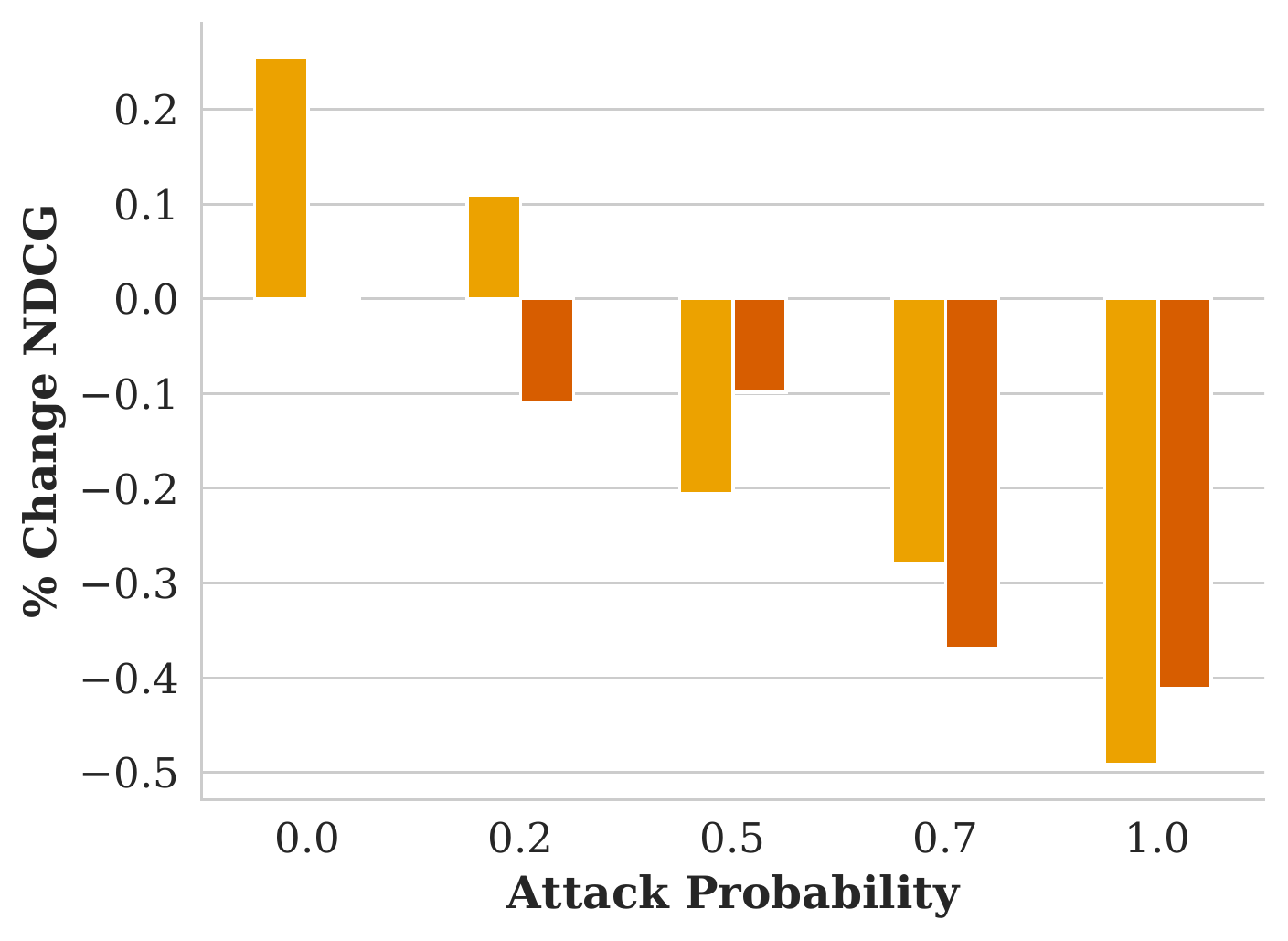}
        \Description{Figure showing how NDCG varies with respect to attack probability $pr$ and fair re-ranking algorithms used by the image search engine.}
        \caption{NDCG}
    \end{subfigure}
    \begin{subfigure}[t]{0.28\textwidth}
    \centering
        \includegraphics[width=\textwidth,keepaspectratio]{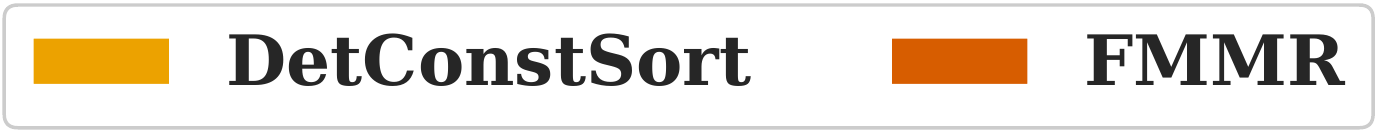}
        \Description{Legend for the above figure, with color coded bars for two fair re-ranking algorithms: DetConstSort and FMMR.}
    \end{subfigure}
    \caption{DetConstSort has poor performance even without an attack, making our results uninteresting.}
    \label{fig:detconstsort}
\end{figure*}

\subsection{Comparison between DetConstSort and FMMR}
\label{sec:detconstsort}

In this section we compare the performance of two fair re-rankers in the presence of our GAP attack. We have already described the details of the first algorithm, FMMR, in \autoref{sec:fmmr}. 

\subsubsection{DetConstSort}

The second algorithm, DetConstSort\cite{geyik2019fairness}, was developed by and is currently deployed at LinkedIn in their talent search system. Unlike FMMR, DetConstSort requires access to the demographic labels of the items it is trying to fairly re-rank. DetConstSort rearranges a given list of items, such that for any particular rank $k$ and for any attribute $a_j$, the attribute is present at least $\lfloor p_{a_j} . k \rfloor$ times in the ranked list, where $p_{a_j}$ is the proportion of items in the list that have the attribute $a_j$. DetConstSort also re-sorts the items within the relevance criteria so that items with better utility scores are placed higher in the ranked list as much as possible, while maintaining the desired attribute ratio. It thus aims to solve a deterministic interval constrained sorting problem.

If ground-truth demographic labels are unavailable, DetConstSort may instead utilize labels sourced from a demographic inference model. Recent work, however, has shown that DetConstSort is sensitive to errors in demographic labels, with one example of such errors being inaccurate inferences~\cite{ghosh2021fair}.

\subsubsection{Evaluation Results}

We present the results of our GAP attacks against our search engine when it uses DetConstSort and FMMR as the fair re-ranker, respectively, in \autoref{fig:detconstsort}. As in \autoref{sec:results}, these results are averaged across three queries, multiple values of $k$, etc.

For DetConstSort, the skew and attention metrics are not impacted by our attack. This can be clearly seen by comparing the $\eta$ values when $pr=0$ (\ie there are no perturbed images) to other values of $pr$: for DetConstSort, $\eta$ for skew and attention starts high (unfair) when $pr=0$, and does not change as $pr$ increases. The correct interpretation of these results is \textbf{not} that DetConstSort is resilient to our attack. Rather, the correct interpretation is that DetConstSort starts off unfair due to the use of inaccurate, inferred demographic data~\cite{ghosh2021fair}, and our attack is unable to make the unfairness worse.

Thus, we find that a prerequisite for evaluating the success of our attacks on DetConstSort is an accurate demographic inference model. Developing such models is still an active area of research, and is out-of-scope for our work. Should a more accurate demographic inference model be designed in the future, however, it must be designed with adversarial robustness in mind to prevent our attacks.

%\cwnote{This needs review by everyone}. The results in \autoref{fig:detconstsort} also highlight an interesting asymmetry. Recall that we train our CGAP models using the same demographic inference models that DetConstSort relies upon for fair re-ranking, and yet the resulting attacks succeed (\eg against FMMR), while the fair re-ranking does not. This disparity suggests that achieving robust, fair ML may be a fundamentally harder problem than compromising those systems with adversarial examples. 
%\todo{So I don't think the attack does not succeed, it's more like the results are already so bad that you can't make eta much worse. Evidence that something is happening is shown by the changes in NDCG, even though they're small. I am not able to say anything more concrete than this since these plots are of mean etas and not any particular list. - Avijit}

\subsection{Choice of Queries}\label{sec:querychoice}

To facilitate our experiments, we chose to select search query terms that would provide a sizeable list of images. To do so, we looked at the list of terms in the COCO image captions (excluding English stop words and words related to ethnicity or gender). The following table shows some top terms. From this information, we composed our three queries given that ``sitting'', ``tennis'', ``table'', ``person'', ``pizza'',  etc. were among the most popular terms.

\begin{table*}[h]
\resizebox{3cm}{!}{%
\centering
\begin{tabular}{lr}
\toprule
Term &      Count \\
\midrule
sitting  &  55084 \\
standing &  44121 \\
people   &  42133 \\
holding  &  29055 \\
large    &  25305 \\
person   &  25123 \\
street   &  21609 \\
table    &  20775 \\
small    &  20661 \\
tennis   &  19718 \\
riding   &  18809 \\
train    &  18287 \\
young    &  17767 \\
red      &  17522 \\
baseball &  15362 \\
pizza    & 11163  \\
\bottomrule
\end{tabular}
}
%\vspace*{9pt}
\caption{The most common (gender or race unrelated) caption terms in the evaluation dataset.}
\label{tab:commonterms}
\end{table*}

\end{document}